\documentclass{article}

\usepackage{microtype}
\usepackage{graphicx}
\usepackage{subcaption}
\usepackage{booktabs}
\usepackage{multirow}
\usepackage{colortbl}
\usepackage[table]{xcolor}
\usepackage{array}
\usepackage{amsmath}
\usepackage{amssymb}
\usepackage{mathtools}
\usepackage{amsthm}
\usepackage[most]{tcolorbox}
\usepackage{hyperref}

\usepackage[preprint]{icml2026}
\usepackage[capitalize,noabbrev]{cleveref}

\makeatletter
\@ifundefined{Highlight}{\definecolor{Highlight}{gray}{0.92}}{}
\@ifundefined{HeaderGray}{\definecolor{HeaderGray}{gray}{0.95}}{}
\definecolor{myblue}{HTML}{1E70B8}
\definecolor{myjujube}{HTML}{990000}
\definecolor{mygray}{gray}{0.45}
\makeatother

\definecolor{mygrey}{gray}{0.6}
\newcolumntype{C}[1]{>{\centering\arraybackslash}p{#1}}
\newcommand{\cc}{\cellcolor{Highlight}}
\newcommand{\realup}[1]{\ensuremath{^{\textcolor{myblue}{\scriptsize\uparrow #1}}}}
\newcommand{\realdown}[1]{\ensuremath{^{\textcolor{myjujube}{\scriptsize\downarrow #1}}}}
\newcommand{\stay}{\ensuremath{^{\makebox[0pt][l]{\textcolor{mygray}{\textbf{--}}}\phantom{\scriptsize\uparrow 0.0}}}}
\newcommand{\dummy}{\ensuremath{^{\phantom{\scriptsize\uparrow 0.0}}}}
\newcommand{\obsbox}[1]{\begin{tcolorbox}[colframe=black!70, colback=lightgray!15, boxrule=1pt, arc=2mm]\small#1\end{tcolorbox}}
\newcommand{\affmark}[1]{\textsuperscript{#1}}
\newcommand{\equalmark}{\textsuperscript{*}}

\theoremstyle{plain}

\theoremstyle{definition}

\theoremstyle{remark}

\icmltitlerunning{See First, Answer Later: Visual Evidence Pre-Alignment via Sufficiency-Driven RL}

\begin{document}

\twocolumn[
  \icmltitle{See First, Answer Later:\\
  Visual Evidence Pre-Alignment via Sufficiency-Driven RL}

  \parbox{\linewidth}{\centering
    {\bfseries
    \begin{tabular}{@{}c@{\hspace{1.1em}}c@{\hspace{1.1em}}c@{\hspace{1.1em}}c@{\hspace{1.1em}}c@{}}
      Yilian Liu\affmark{1}\equalmark &
      Sicong Leng\affmark{2}\equalmark &
      Guoshun Nan\affmark{1} &
      Junyi Zhu\affmark{1} &
      Jiayu Huang\affmark{1}
    \end{tabular}\\
    \begin{tabular}{@{}c@{\hspace{1.1em}}c@{\hspace{1.1em}}c@{\hspace{1.1em}}c@{\hspace{1.1em}}c@{}}
      Minghao Sun\affmark{1} &
      Xuancheng Zhu\affmark{1} &
      Yisong Chen\affmark{3} &
      Zexian Wei\affmark{1} &
      Xiaofeng Tao\affmark{1}
    \end{tabular}
    }\\[0.8ex]
    {\normalfont
    \affmark{1}Beijing University of Posts and Telecommunications, China\\[0.8ex]
    \affmark{2}Nanyang Technological University, Singapore; \space
    \affmark{3}China Telecom, China\\[0.5em]
    {\small\texttt{\{liuyilian,nanguo2021\}@bupt.edu.cn; Lengsicong@gmail.com}}\\[0.4ex]
    
    }
  }

  \icmlcorrespondingauthor{Guoshun Nan}{nanguo2021@bupt.edu.cn}


  \vskip 0.1in
]
\printAffiliationsAndNotice{
    \textsuperscript{*} Equal contribution
  }

\begin{abstract}
  Multimodal large language models (MLLMs) integrate strong text reasoning with visual inputs, yet their responses can be inconsistent with the underlying images, indicating ineffective utilization of visual evidence during inference.
The prevailing training paradigm relies on large-scale caption-based pretraining for general alignment, followed by supervised fine-tuning and reinforcement learning to enable instruction following and complex reasoning.
However, such pretraining provides only weak visual grounding: short, coarse captions bias models toward salient objects while neglecting fine-grained visual evidence.
In this paper, we introduce \textbf{V}isual \textbf{E}vidence \textbf{P}re-\textbf{A}lignment (\textbf{VEPA}), an intermediate stage between pretraining and post-training that explores a novel sufficiency-driven objective with Group Relative Policy Optimization (GRPO) to optimize question-conditioned visual evidence descriptions.
Extensive experiments across diverse benchmarks show that our VEPA consistently enhances performance on visually demanding evaluations and complements standard supervised post-training.
Further analyses show that the income stems from strengthened, transferable visual grounding, rather than from additional task-specific training.
\end{abstract}

\section{Introduction}
\label{sec:intro}
\begin{figure}[ht]

  \begin{center}
    \centerline{\includegraphics[width=\columnwidth]{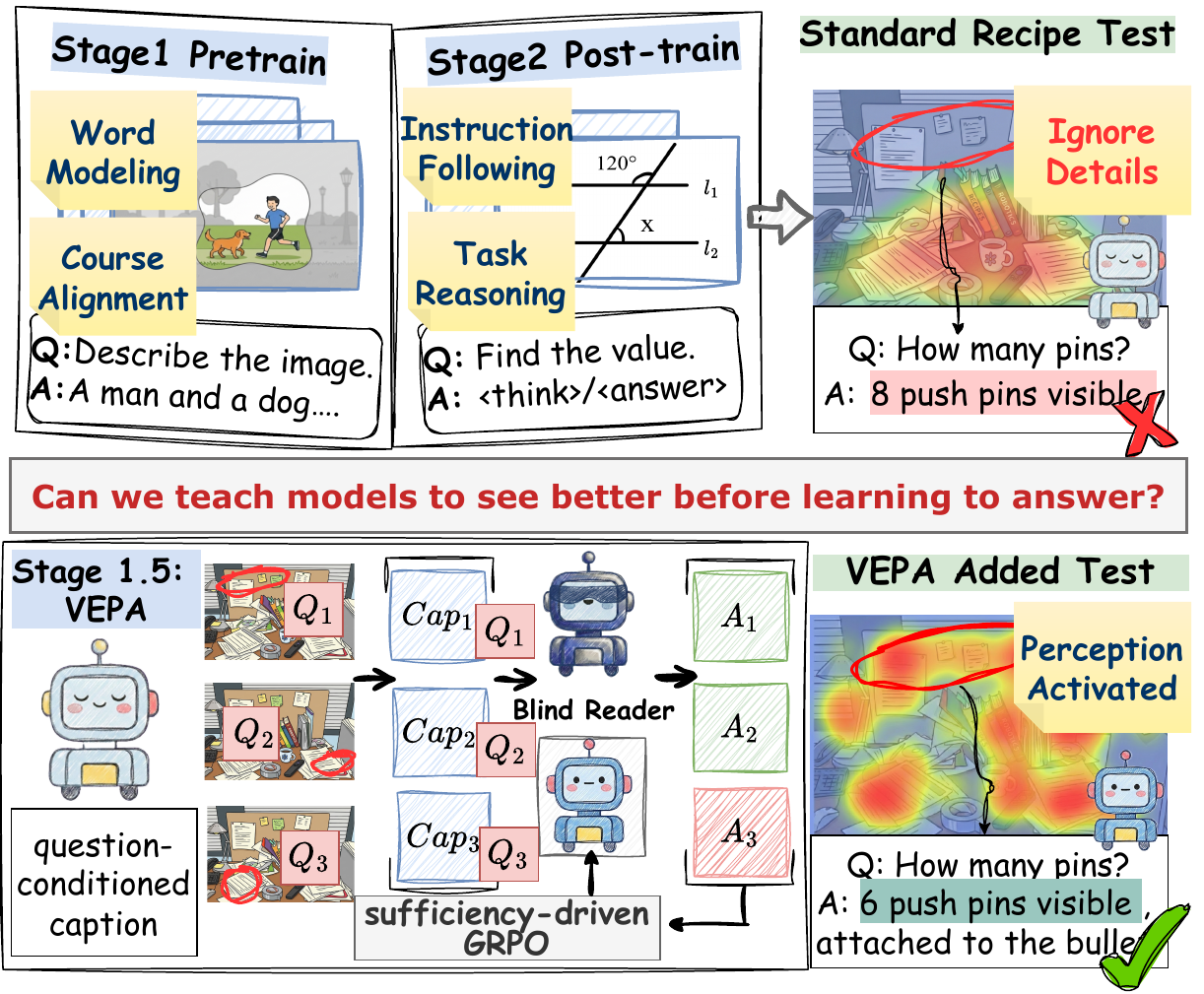}}
\caption{\textbf{Motivation and overview.} The standard two-stage recipe often yields coarse alignment and encourages shortcut answering that ignores visual details. We insert VEPA as an intermediate stage that trains the model to produce question-conditioned visual evidence using \textbf{sufficiency-driven} GRPO, with a frozen blind reader (an LLM) answering and verifying whether the evidence suffices to recover the answer. This ``see first, answer later'' pre-alignment activates perceptual ability and improves downstream visual grounding.}
    \label{fig:intro}
  \end{center}
  \vskip -0.4in
\end{figure}

Multimodal large language models (MLLMs) have recently achieved strong performance on diverse tasks such as document and chart comprehension and diagrammatic math reasoning.
Despite these advances, MLLMs may produce outputs that are weakly grounded in the underlying image~\cite{luoprobing,xia2025visionary}, including omissions of critical visual details, incorrect attribution of attributes or relations, and hallucinated content~\cite{li2025visual, xia2025visionary}.
Such failures are not confined to specific tasks or domains but instead reflect a pervasive limitation in how visual evidence is exploited during inference.
Most existing MLLM training pipelines follow a two-stage paradigm~\cite{liu2023visual, zhu2024minigpt}.
In the pretraining stage, models are trained on large-scale image–caption corpora to establish coarse vision–language alignment~\cite{li2020oscar}.
In the post-training stage, supervised fine-tuning, often combined with reinforcement learning, is applied to improve instruction following and downstream task performance~\cite{christiano2017deep,zhai2024fine}.
However, caption-driven pretraining alone is insufficient to ensure robust visual perception prior to post-training.
Captions are typically short and coarse, emphasizing salient objects~\cite{lin2014microsoft} or global scene descriptions while omitting many detailed attributes, relations, and less prominent regions.
This supervision biases models toward a narrow subset of visual content and provides limited incentive to encode fine-grained or question-relevant visual evidence, leading models to rely disproportionately on language priors during inference~\cite{goyal2017making}.

A straightforward remedy is to strengthen pretraining supervision by replacing short captions with dense descriptions~\cite{zheng2024dreamlip, zeng2025enhancing} or augmenting training with perception-oriented datasets (e.g., OCR) that explicitly encode fine-grained visual content.
However, this approach is limited both practically and fundamentally.
High-quality dense captions and OCR~\cite{chen2024scaling} annotations are costly~\cite{dong-etal-2025-scalable, liu2025visual, shen2025vlm} to collect at scale for diverse web images, and manual annotation pipelines introduce additional biases, omissions, and inconsistencies~\cite{misra2016seeing, hu2023promptcap}.
More fundamentally, the high information density of visual inputs compared to textual representations makes it inherently difficult for any finite description to faithfully encode all objects, relations, and spatial details in a scene.
Thus, even dense captions remain a lossy~\cite{dubois2021lossy} and biased proxy for visual content~\cite{chen2024makes}, and simply scaling static image–text pairs is insufficient to achieve the level of perceptual precision required by complex multimodal reasoning tasks.

In this work, we introduce \textbf{V}isual \textbf{E}vidence \textbf{P}re-\textbf{A}lignment (\textbf{VEPA}), an intermediate training stage that teaches MLLMs to generate question-conditioned visual evidence, textual descriptions that captures the image details needed to solve a given question.
As illustrated in Figure~\ref{fig:intro}, VEPA optimizes evidence generation via sufficiency-driven Group Relative Policy Optimization (GRPO)~\cite{shao2024deepseekmath}.
During training, a frozen blind reader (an LLM), conditioned only on the question and the generated evidence, serves as an auxiliary evaluator by verifying whether the evidence is sufficient to recover the ground-truth answer.
We carefully design the reward to discourage answer leakage and degenerate repetitive evidence, thereby decoupling visual grounding from answer generation.
Across diverse evaluation settings, VEPA consistently outperforms the standard training recipe on benchmarks spanning knowledge-intensive and compositional VQA, fine-grained perception, and holistic multimodal evaluation. 
Ablation studies and qualitative analyses show that these gains arise from strengthened visual perception rather than additional task-level supervision, and that the learned grounding transfers robustly to out-of-distribution data.

In summary, our contributions are in three aspects:
\begin{itemize}
\item We propose \textbf{V}isual \textbf{E}vidence \textbf{P}re-\textbf{A}lignment (\textbf{VEPA}), a novel intermediate training stage between pretraining and post-training that explicitly strengthens visual perception prior to task-level instruction tuning, encouraging MLLMs to attend to and encode relevant visual evidence before answering.

\item We instantiate VEPA with a GRPO-based reinforcement learning framework that trains models to generate question-conditioned visual evidence.
A frozen blind reader evaluates whether the generated evidence suffices to recover the ground-truth answer, decoupling visual grounding from answer generation and enabling training with existing QA data without additional annotation cost.

\item Extensive experiments across diverse benchmarks and model backbones demonstrate consistent improvements on visually demanding evaluations.
Ablation studies and qualitative analyses further show that VEPA strengthens visual grounding and induces transferable perceptual capabilities that generalize to out-of-distribution settings.
\end{itemize}

\section{Related Work}

\textbf{Multimodal pretraining and post-training.}
MLLMs follow a two-stage recipe: large-scale vision--language pretraining for coarse alignment, followed by instruction tuning to improve task-level behaviors~\cite{alayrac2022flamingo, li2023blip, liu2023visual, liu2024improved, wang2024qwen2}.
This paradigm has produced strong systems across late-fusion architectures that connect a frozen or lightly-tuned vision encoder to an LLM as well as increasingly capable end-to-end or early-fusion variants.
Recent backbones such as Qwen2-VL~\cite{wang2024qwen2} further improve resolution handling and general visual understanding, strengthening the foundation for downstream multimodal adaptation~\cite{wang2024qwen2}. However, empirical evaluations consistently suggest that coarse caption-driven supervision alone does not always yield reliable fine-grained grounding required by complex VQA-style queries~\cite{fu2025mme}.

\textbf{Visual grounding.}
A recurring challenge for MLLMs is the tendency to under-use visual input, leading to omissions or hallucinated content~\cite{li2025visual, xia2025visionary} when language priors~\cite{goyal2017making} dominate generation.
This issue has motivated dedicated benchmarks and diagnostics that quantify visual faithfulness and robustness beyond raw accuracy, including object-hallucination probes and broad-spectrum evaluation suites~\cite{li2023evaluating, rohrbach2018object}.
Complementary lines of work mitigate hallucinations via improved post-training objectives~\cite{zheng2025deepeyes}, decoding-time constraints~\cite{li2025visual}, or auxiliary verification mechanisms that encourage consistency with perceived evidence~\cite{zheng2025deepeyes}. Despite these advances, many approaches still optimize answer generation directly~\cite{wang2025vl}, making it difficult to isolate and strengthen the upstream visual extraction process that should support downstream reasoning.

\textbf{Intermediate evidence and verifier-based reinforcement learning.}
Generating intermediate representations—such as rationales, descriptions, or other textual evidence—has been explored to improve grounding and interpretability in multimodal reasoning, where models must surface query-relevant visual details~\cite{li2018vqa,rao2021first}.
When reliable intermediate supervision is unavailable, recent language-model post-training leverages verifier-style or preference-based signals as scalable supervision without token-level labels~\cite{rafailov2023direct, wen2025reinforcement}.
In particular, group-based policy optimization such as GRPO removes an explicit critic and estimates baselines from within-group scores, simplifying RL fine-tuning for long responses~\cite{shao2024deepseekmath}.
Our work aligns with this verifier-based direction but targets a different object.
Rather than optimizing the final answer directly, we train a question-conditioned evidence channel to be sufficient for solving and resistant to shortcut behaviors, bridging coarse pretraining and downstream post-training.

\section{Preliminary}

\subsection{Evidence Decomposition for Visual Grounding}
\label{subsec:Evidence decomposition for visual grounding}

The goal of an MLLM is to generate an answer $a$ conditioned on an image $v$ and a question $q$, ideally approximating $P(a \mid v, q)$.
In practice, caption-based pretraining provides only coarse vision–language alignment, and answer-level supervision makes improvements in perception mostly a byproduct of reasoning learning.
As a result, models may over-rely on language priors, drifting toward $P(a \mid q)$ and under-utilizing the visual input $v$.

To address this, we decouple visual perception from reasoning by introducing a latent variable $e$, termed \emph{visual evidence}, which serves as an information bottleneck between perception and answer generation.
The generation process is formalized as:
\begin{equation}
    P(a \mid v, q) = \sum_{e \in \mathcal{E}} P(a \mid e, q) \cdot P(e \mid v, q).
\end{equation}
Here, $P(e \mid v, q)$ acts as a visual representation policy (MLLM) that maps the image and question to a textual evidence $e$, and $P(a \mid e, q)$ is a reasoning policy (LLM) that produces the answer from the evidence and the question.

Under this decomposition, improving visual alignment amounts to shaping $P(e \mid v, q)$ such that the induced evidence is both informative for answering and genuinely grounded in the image.
We therefore require the evidence $e$ to satisfy two properties:

\noindent\textbf{Sufficiency.} Conditioned on the question, the evidence should contain enough information to determine the answer:
\begin{equation}
   P(a \mid v, q)
   \approx
   P(a \mid e, q)
   \quad \text{for } e \sim P(e \mid v, q).
   \label{eq:sufficiency}
\end{equation}
Once $e$ is known, the residual contribution of the raw image $v$ to predicting $a$ should be small.

\noindent\textbf{Visual dependence.} The evidence must depend on both the image and the question, rather than being reconstructible from either alone.
In particular, for image--question pairs $(v_1, q)$ and $(v_2, q)$, or $(v, q_1)$ and $(v, q_2)$, that induce different answers, we require:
\begin{equation}
    P(e \mid v_1, q) \neq P(e \mid v_2, q),
    P(e \mid v, q_1) \neq P(e \mid v, q_2).
    \label{eq:visual-dependence}
\end{equation}

When the above properties hold approximately, the model cannot satisfy the objective by ignoring the image.
Instead, it must route visual information through the evidence channel $P(e \mid v, q)$ before producing an answer.
The intermediate training stage we introduce is designed to explicitly encourage these properties.

\subsection{Perception Coverage from Data Diversity}

Let $\mathcal{D}$ denote the training distribution over triplets $(v, q, a)$.
While the sufficiency and dependence conditions above are defined for individual instances, our objective is to enforce them approximately on average over $\mathcal{D}$.

Concretely, we would like the evidence policy $P(e \mid v, q)$ to satisfy:
\begin{equation}
    \mathbb{E}_{(v, q, a) \sim \mathcal{D}}
    \Big[
        \mathrm{KL}\big(P(a \mid v, q)\,\big\|\,P(a \mid e, q)\big)
    \Big]
    \le \varepsilon_{\mathrm{suff}},
    \label{eq:dataset-suff}
\end{equation}
for a small $\varepsilon_{\mathrm{suff}} \ge 0$, ensuring that answers can be predicted nearly as well from $(e, q)$ as from $(v, q)$ on average.
In addition, to prevent degeneracy, the evidence must retain non-trivial information about both the image and the question, for example,
\begin{equation}
    I_{\mathcal{D}}(e; v \mid q) \ge \delta_v,
    \qquad
    I_{\mathcal{D}}(e; q \mid v) \ge \delta_q,
    \label{eq:dataset-dependence}
\end{equation}
with $\delta_v, \delta_q > 0$, where $I_{\mathcal{D}}(\cdot;\cdot\mid\cdot)$ denotes conditional mutual information under the joint distribution induced by $\mathcal{D}$ and $P(e \mid v, q)$.

Rather than supervising $e$ with increasingly dense captions, which remain fundamentally constrained by the information density of text, we leverage the diversity inherent in VQA-style data.
When triplets $(v, q, a)$ are sampled from a heterogeneous corpus, different questions probe distinct objects, attributes, relations, and regions within the same image, including details rarely emphasized in caption-based supervision.
\textbf{Encouraging sufficiency and dependence over $\mathcal{D}$ therefore pushes the evidence policy to cover a broad range of task-relevant visual details. }
Each evidence instance need only be sufficient for its own question, while the collection of question--evidence pairs across the dataset provides complementary and diverse textual views of visual content.

\begin{figure*}[!tbp]

  \begin{center}
    \centerline{\includegraphics[width=0.9\textwidth]{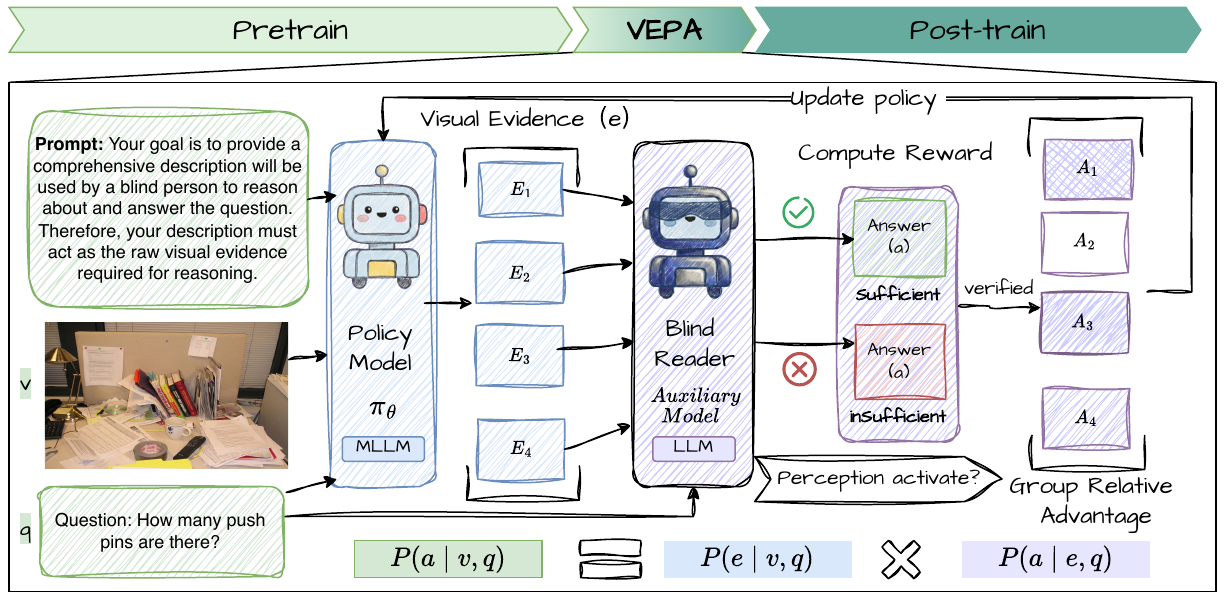}}
\caption{\textbf{Framework of VEPA.} VEPA is inserted between pretraining and post-training. Given an image $v$ and a question $q$, the policy MLLM $\pi_\theta$ is prompted to generate question-conditioned \emph{visual evidence} and samples a group of candidate visual evidence $\{e\}$. A frozen text-only blind reader (auxiliary LLM) answers using only $(q, e)$. We optimize
$\pi_\theta$ with a novel sufficiency-driven objective via GRPO.}
    \vskip -0.3in
    \label{fig:method}
  \end{center}
\end{figure*}
\section{Methodology}
\label{sec:method}

\subsection{Evidence Generation as Perceptual Pre-Alignment}

We propose \textbf{Visual Evidence Pre-Alignment (VEPA)}, an intermediate training stage that explicitly aligns visual perception before task-level post-training.
As illustrated in Figure~\ref{fig:method}, VEPA trains an MLLM to generate \emph{question-conditioned visual evidence}, textual descriptions that encode the visual information for a given question, without directly optimizing answer generation.
This section first formalizes the evidence-generation objective (\S\ref{subsec:evidence-objective}), then presents the VEPA optimization framework (\S\ref{subsec:vepa-optimization}), and finally describes the blind-reader-based reward design (\S\ref{subsec:blind-reader-reward}).

\label{subsec:evidence-objective}

We instantiate the evidence decomposition introduced in \S\ref{subsec:Evidence decomposition for visual grounding} as an intermediate training objective.
Since ground-truth visual evidence annotations are unavailable at scale and dense captions remain both costly and incomplete, supervised learning over evidence tokens is infeasible.
So we cast evidence generation as a policy optimization problem.

Let $\mathcal{D}=\{(v_i,q_i,a_i)\}$ denote a VQA training set.
During VEPA, the MLLM is not trained to produce the final answer $a$.
Instead, for each triplet $(v,q,a)$, it generates an intermediate textual sequence $e$ intended to encode the visual information necessary to answer $q$.
For each rollout, the policy samples a single evidence sequence up to a fixed length.
Learning is driven solely by a sequence-level reward computed after the evidence is fully generated.

\vspace{-0.8em}
\subsection{VEPA Optimization Framework}
\label{subsec:vepa-optimization}

The proposed VEPA optimizes the evidence policy $\pi_\theta(e\mid v,q)$ using reinforcement learning with a sequence-level reward $R(e;v,q,a^*)$.
Given a dataset $\mathcal{D}$ of triplets $(v,q,a^*)$, the objective is
\begin{equation}
    J(\theta)
    =
    \mathbb{E}_{(v,q,a^*)\sim\mathcal{D}}
    \mathbb{E}_{e\sim\pi_\theta(\cdot\mid v,q)}
    \bigl[R(e;v,q,a^*)\bigr].
    \label{eq:rl-objective}
\end{equation}

Optimizing this objective directly is unstable for long, free-form evidence sequences.
We therefore adopt sufficiency-driven \emph{Group Relative Policy Optimization (GRPO)} as a practical optimization strategy\footnote{Importantly, GRPO serves solely as an optimizer for VEPA’s evidence objective; our contribution lies in the formulation of the objective and reward, not in the optimization algorithm itself.}.

At each update step, for a given triplet $(v,q,a^*)$, we sample a group of $G$ candidate evidence sequences $\{e_g\}_{g=1}^G$ from the current policy.
Each candidate receives a scalar reward $R_g$, from which we compute a group-relative standardized advantage
\begin{equation}
    A_g=\frac{R_g-\bar{R}}{\sigma_R+\delta},
\end{equation}
where $\bar{R}$ and $\sigma_R$ denote the group mean and standard deviation.
Policy updates follow a PPO-style clipped surrogate objective with a KL regularization~\cite{schulman2017proximal} term that constrains deviation from a frozen reference policy $\pi_{\mathrm{ref}}$ (the pretrained model before VEPA):
\begin{equation}
\begin{aligned}
J_{\mathrm{VEPA}}(\theta)
&=
\mathbb{E}
\Bigg[
\frac{1}{G}
\sum_{g=1}^G
\frac{1}{T_g}
\sum_{t=1}^{T_g}
\mathcal{L}^{\mathrm{clip}}_{g,t}(\theta)
\\
&\qquad\quad
-
\beta\,
\mathbb{D}_{\mathrm{KL}}\!\left(
\pi_\theta \,\|\, \pi_{\mathrm{ref}}
\right)_t
\Bigg].
\end{aligned}
\end{equation}
This formulation enables stable optimization of long evidence sequences while preserving linguistic coherence.

\subsection{Blind Reader-Based Sufficiency Reward}
\label{subsec:blind-reader-reward}

To operationalize the sufficiency criterion in \S\ref{subsec:Evidence decomposition for visual grounding} without evidence annotations, we introduce a frozen \emph{blind reader} model $f_\phi$.
The blind reader is an instruction-tuned language model that observes only the generated evidence $e$ and question $q$, and never accesses the image.

For each training example, the blind reader is prompted to answer the question using the evidence alone and to indicate whether the evidence directly leaks the answer.
Let $\hat{a}(e,q)$ denote the predicted answer after normalization.
Based on this answer, we then define a binary solvability score
\begin{equation}
    s(e,q,a^*)=\mathbb{I}[\hat{a}(e,q)=a^*],
\end{equation}
which encourages the evidence to be sufficient for answers.

To discourage trivial solutions that restate the answer, the blind reader emits an honesty flag indicating whether answer leakage is detected.
We define
\begin{equation}
    h(e,q)\in\{0,1\},
\end{equation}
where $h=1$ denotes honest evidence.

Finally, to suppress degenerate repetition, we apply a weak penalty $\gamma(e)\in(0,1]$ based on simple repetition statistics.
The resulting sequence-level reward can be expressed as
\begin{equation}
    R(e,q,a^*) = s(e,q,a^*) \cdot h(e,q) \cdot \gamma(e).
\end{equation}

Maximizing the expected reward encourages evidence that is sufficient for a blind reader while remaining image-grounded and non-degenerate, without requiring any human evidence supervision.

Overall, the proposed VEPA reframes visual grounding as an explicit pre-alignment problem by isolating evidence generation from answer prediction.
By optimizing question-conditioned visual evidence using a blind-reader–based sufficiency signal, VEPA encourages models to encode task-relevant visual information before engaging in downstream reasoning.
This design enables effective perceptual alignment without additional annotations and is complementary to standard supervised post-training, providing a principled mechanism for improving visual grounding in MLLMs.

\section{Experiments}
\label{sec:experiments}
We conduct experiments to evaluate the effectiveness of our VEPA, and focus on the following questions:

\textbf{RQ1. Downstream Performance.}
Does VEPA consistently improve performance across diverse multimodal benchmarks compared to the standard training pipeline? \\
\textbf{RQ2. Visual Grounding.}
Does VEPA strengthen visual grounding by encouraging the generation of question-conditioned evidence that is sufficient for reasoning? \\
\textbf{RQ3. Visual Reliance.}
Does VEPA increase reliance on visual inputs over language priors, as reflected by robustness to visual perturbations and reduced hallucination? \\
\textbf{RQ4. Training Dynamics.}
How does visual perception evolve over the course of VEPA training?

\subsection{Experimental Setup}
\textbf{Implementation details.}
We use Qwen2-VL-2B~\cite{wang2024qwen2} as the backbone for all model variants.
During the VEPA stage, we optimize the evidence-generation policy using our sufficiency-driven GRPO on a curated subset of FineVision~\cite{wiedmann2025finevision}.
Specifically, we uniformly sample 5k training instances from the ScienceQA~\cite{lu2022learn}, AI2D-Merge~\cite{kembhavi2016diagram}, ChartQA~\cite{masry2022chartqa}, Geo3K~\cite{lu2021inter}, TextVQA~\cite{singh2019towards}, and CLEVR~\cite{johnson2017clevr}.
These datasets jointly cover complementary perception-centric skills, including diagram and scientific reasoning, chart and plot understanding, geographic and map recognition, text understanding in natural images, and compositional visual reasoning.

To better align with the objective of visually grounded evidence generation, we further prioritize questions with high visual dependency, such that the correct answer cannot be reliably inferred from the question alone.
This filtering yields diverse evidence patterns that better approximate the target distribution and improves the robustness of the learned evidence policy.

All GRPO training is implemented using \textsc{VeRL}.
To compute the evidence-sufficiency reward, we employ a frozen text-only blind reader based on Qwen2.5-7B-Instruct~\cite{qwen2025qwen25technicalreport}, which predicts answers conditioned solely on the question and the generated evidence.
Unless otherwise specified, we report exact-match accuracy on downstream VQA benchmarks with identical decoding settings across all variants.
Additional implementation details and hyperparameters are provided in Appendix~\ref{appdix:experiment details}.

\subsection{RQ1: Downstream Performance Across Tasks}

To assess whether inserting VEPA between pretraining and standard post-training improves downstream performance, we conduct controlled comparisons in which all variants share the same backbone, supervised fine-tuning (SFT) data, and SFT hyperparameters.
Specifically, all models are initialized from Qwen2-VL-2B, trained with standard SFT on 20k examples, and evaluated either directly (SFT) or after an additional VEPA Etage followed by SFT (\textbf{VEPA+SFT}).

\begin{table*}[t]
\centering
\footnotesize
\renewcommand{\arraystretch}{0.9}

\caption{\textbf{Downstream benchmark results under different SFT data settings.}
Accuracy (\%) is reported for each target benchmark.
Rows are grouped by the SFT dataset used for post-training (\textbf{SFT data}), comparing the SFT baseline against \textbf{VEPA+SFT} (shaded).
\textbf{Avg.} denotes the macro-average over all benchmarks.
$^{\dagger}$ marks \emph{in-domain} evaluation.
Subscripts indicate the change relative to SFT (\textcolor{myblue}{$\uparrow$} improvement, \textcolor{myjujube}{$\downarrow$} decline, \textcolor{mygray}{-} no change).}

\label{tab:rq1_main}

\resizebox{\linewidth}{!}{
    \begin{tabular}{ll ccccccc}
    \toprule
    \rowcolor{HeaderGray}
    \textbf{SFT data} & \textbf{Method} & \textbf{A-OKVQA} & \textbf{ChartQA} & \textbf{TextVQA} & \textbf{GQA} & \textbf{MME} & \textbf{MMStar} & \textbf{Avg.} \\
    \midrule

    \multirow{2}{*}{A-OKVQA}
     & SFT & 59.70$^{\dagger}$\dummy & 51.88\dummy & 62.73\dummy & 53.70\dummy & 75.22\dummy & 41.07\dummy & 57.38\dummy \\
     & \cc \textbf{VEPA+SFT}
     & \cc \textbf{66.60}$^{\dagger}$\realup{6.9}
     & \cc \textbf{58.96}\realup{7.1}
     & \cc \textbf{66.78}\realup{4.1}
     & \cc \textbf{54.75}\realup{1.1}
     & \cc \textbf{76.95}\realup{1.7}
     & \cc \textbf{42.80}\realup{1.7}
     & \cc \textbf{61.14}\realup{3.8} \\
    \midrule

    \multirow{2}{*}{ChartQA}
     & SFT & \textbf{59.60}\dummy & 69.44$^{\dagger}$\dummy & 77.56\dummy & 59.70\dummy & 81.70\dummy & 40.93\dummy & 64.82\dummy \\
     & \cc \textbf{VEPA+SFT}
     & \cc \textbf{59.60}\stay
     & \cc \textbf{70.28}$^{\dagger}$\realup{0.8}
     & \cc \textbf{78.96}\realup{1.4}
     & \cc \textbf{60.99}\realup{1.3}
     & \cc \textbf{82.13}\realup{0.4}
     & \cc \textbf{41.93}\realup{1.0}
     & \cc \textbf{65.65}\realup{0.8} \\
    \midrule

    \multirow{2}{*}{TextVQA}
     & SFT & 34.95\dummy & 68.08\dummy & 80.66$^{\dagger}$\dummy & 38.40\dummy & \textbf{82.13}\dummy & 41.87\dummy & 57.68\dummy \\
     & \cc \textbf{VEPA+SFT}
     & \cc \textbf{35.50}\realup{0.6}
     & \cc \textbf{68.64}\realup{0.6}
     & \cc \textbf{81.44}$^{\dagger}$\realup{0.8}
     & \cc \textbf{40.60}\realup{2.2}
     & \cc 80.55\realdown{1.6}
     & \cc \textbf{42.40}\realup{0.5}
     & \cc \textbf{58.19}\realup{0.5} \\
    \midrule

    \multirow{2}{*}{GQA}
     & SFT & 42.00\dummy & 60.40\dummy & 63.49\dummy & 64.41$^{\dagger}$\dummy & 78.67\dummy & 43.07\dummy & 58.67\dummy \\
     & \cc \textbf{VEPA+SFT}
     & \cc \textbf{42.10}\realup{0.1}
     & \cc \textbf{64.40}\realup{4.0}
     & \cc \textbf{66.15}\realup{2.7}
     & \cc \textbf{64.52}$^{\dagger}$\realup{0.1}
     & \cc \textbf{80.55}\realup{1.9}
     & \cc \textbf{43.80}\realup{0.7}
     & \cc \textbf{60.25}\realup{1.6} \\
    \midrule

    \multirow{2}{*}{Mixed}
     & SFT & 62.50$^{\dagger}$\dummy & 70.32$^{\dagger}$\dummy & 79.73$^{\dagger}$\dummy & 51.07$^{\dagger}$\dummy & 79.97\dummy & 44.53\dummy & 64.69\dummy \\
     & \cc \textbf{VEPA+SFT}
     & \cc \textbf{62.70}$^{\dagger}$\realup{0.2}
     & \cc \textbf{70.44}$^{\dagger}$\realup{0.1}
     & \cc \textbf{79.88}$^{\dagger}$\realup{0.2}
     & \cc \textbf{53.69}$^{\dagger}$\realup{2.6}
     & \cc \textbf{80.26}\realup{0.3}
     & \cc \textbf{45.07}\realup{0.5}
     & \cc \textbf{65.34}\realup{0.7} \\
    \bottomrule
    \end{tabular}
}
\vspace{-1em}
\end{table*}

We consider four single-domain SFT settings using A-OKVQA~\cite{schwenk2022okvqa}, ChartQA~\cite{masry2022chartqa}, TextVQA~\cite{singh2019towards}, and GQA~\cite{hudson2019gqa}, as well as a \emph{Mixed} setting constructed by uniformly sampling 5k examples from each dataset.
Models are evaluated on six benchmarks, including in-domain evaluations (marked with $^{\dagger}$), out-of-domain transfer benchmarks, and two general-purpose multimodal benchmarks (MME and MMStar).
Table~\ref{tab:rq1_main} reports accuracy comparisons.

\noindent\textbf{Overall performance trends.}
Across all five SFT settings, VEPA consistently improves average performance (+0.5 to +3.8 points) without degrading in-domain accuracy.
In-domain results are either preserved or modestly improved, indicating that VEPA does not trade task specialization for robustness.
This suggests that the VEPA stage provides complementary supervision signals that are compatible with downstream SFT objectives.

\noindent\textbf{Generalization under domain shift.}
Performance gains are often more pronounced on out-of-domain benchmarks than on the in-domain metric.
In particular, perception-intensive tasks such as ChartQA and TextVQA frequently exhibit larger improvements.
For example, under A-OKVQA SFT, VEPA substantially improves ChartQA and TextVQA while also improving A-OKVQA itself; under GQA SFT, VEPA again yields sizable gains on ChartQA and TextVQA, with comparatively small changes on in-domain GQA.
Such asymmetric improvements are unlikely to arise from generic additional training, which would be expected to affect all benchmarks more uniformly.
Instead, they are consistent with VEPA strengthening transferable, question-conditioned visual grounding.

\noindent\textbf{Complementarity with data diversification.}
If VEPA merely compensated for limited domain coverage in SFT, its effect should diminish under the Mixed SFT setting.
However, VEPA continues to improve average performance and yields a notable gain on GQA even with a stronger, diversified SFT baseline.
This indicates that VEPA enforces a distinct inductive bias, greater reliance on visual evidence, rather than functioning solely as a substitute for broader supervised data.

\noindent\textbf{General multimodal evaluation.}
VEPA also improves performance on MME and MMStar across SFT settings.
Since these benchmarks are not aligned with any specific SFT domain, the gains suggest that VEPA enhances the model’s general ability to ground answers in visual evidence, rather than improving benchmark-specific patterns.

Overall, these results demonstrate that VEPA serves as an effective intermediate alignment stage: it improves average downstream performance, preserves in-domain accuracy, and yields the largest benefits under domain shift, where transferable perceptual grounding is most critical.

\obsbox{\textbf{Insight~I.} VEPA consistently improves downstream performance and robustness by activating transferable visual grounding beyond standard supervised fine-tuning.}

\subsection{RQ2: Evidence Sufficiency and Selectivity}
\label{subsec:rq2}

While RQ1 establishes that VEPA improves downstream performance, it does not directly verify whether VEPA optimizes the intended objective—namely, producing question-conditioned visual evidence that is sufficient for reasoning.
RQ2 therefore evaluates the quality of the learned evidence representations.
\begin{table}[h]
  \caption{\textbf{Blind-reader evaluation of evidence sufficiency.}
  We prompt the base model and the model after the VEPA stage to generate question-conditioned descriptions, and provide only the question and the generated description to a frozen text-only blind reader (Qwen2.5-7B-Instruct) to answer POPE and MMStar without image access.
  We report Acc. (\%, accuracy) and the average description length (tokens).}
  \label{tab:base-vs-grpo}
  \centering
  \small
  \renewcommand{\arraystretch}{1.05}
  \setlength{\tabcolsep}{6pt}

  \begin{tabular}{lcc@{\hspace{10pt}}cc}
    \toprule
    \multirow{2}{*}{\textbf{Method}} & \multicolumn{2}{c}{\textbf{MMStar}} & \multicolumn{2}{c}{\textbf{POPE}} \\
    \cmidrule(lr){2-3}\cmidrule(lr){4-5}
     & \textbf{Acc.} & \textbf{Length} & \textbf{Acc.} & \textbf{Length} \\
    \midrule
    Base & 42.53 & 201.97 & 78.05 & 464.21 \\
    \textbf{VEPA} & \textbf{44.27} & \textbf{187.86} & \textbf{79.25} & \textbf{365.16} \\
    \bottomrule
  \end{tabular}

  \vskip -0.1in
\end{table}
For each image–question pair, we prompt either the base model or the model after the VEPA stage to generate a question-conditioned visual description, explicitly discouraging direct answer disclosure.
We then provide only the question and the generated description to a frozen text-only blind reader (Qwen2.5-7B-Instruct), which attempts to answer POPE and MMStar without access to the image.
Because the blind reader never observes visual inputs, higher accuracy directly reflects whether the generated evidence surfaces task-relevant visual information.

Table~\ref{tab:base-vs-grpo} shows that VEPA consistently improves blind-reader accuracy on both benchmarks, from 42.53\% to 44.27\% on MMStar and from 78.05\% to 79.25\% on POPE.
Crucially, these gains are accompanied by \emph{shorter} descriptions: the average output length decreases from 201.97 to 187.86 tokens on MMStar and from 464.21 to 365.16 tokens on POPE.
This rules out verbosity as a trivial explanation for the improved solvability.

Consistent with this observation, the mean rollout length exhibits an initial transient increase followed by a gradual decrease and stabilization, rather than unbounded growth (See details in Appendix~\ref{appendix:training}).
After an initial transient increase, it rapidly stabilizes, indicating convergence toward concise yet informative evidence rather than progressively longer descriptions.
Taken together, these results indicate that VEPA improves the \emph{selectivity and sufficiency} of generated evidence, enabling task-critical visual cues to be externalized in a compact form.
This provides a direct mechanistic explanation for the stable downstream gains observed in RQ1.

\obsbox{\textbf{Insight~II.} VEPA promotes concise yet sufficient visual evidence, making task-relevant visual information recoverable without relying on verbose descriptions.}

\subsection{RQ3: Reliance on Visual Inputs}

\begin{table}[h]
\centering
\caption{\textbf{Robustness under image corruption.}
Models are trained with A-OKVQA SFT.
We report clean accuracy (\%), retention under three corruption types
(Blur, Partial noise, Pure noise), and AUC of the retention curve (lower is better).}
\label{tab:rq3_corruption}
\footnotesize
\renewcommand{\arraystretch}{1.10}
\setlength{\tabcolsep}{4pt}

\begin{tabular}{lcccc}
\toprule
\multirow{2}{*}[-0.6ex]{\textbf{Metric}} &
\multicolumn{2}{c}{\textbf{ChartQA}} &
\multicolumn{2}{c}{\textbf{GQA}} \\
\cmidrule(lr){2-3}\cmidrule(lr){4-5}
& \textbf{SFT} & \textbf{VEPA+SFT}
& \textbf{SFT} & \textbf{VEPA+SFT} \\
\midrule
Clean Acc. (\%)             & 51.88 & \textbf{58.96} & 52.13 & \textbf{53.07} \\
Blur ($r$)                  & 0.084 & 0.074 & 0.838 & 0.836 \\
Partial ($r$)               & 0.753 & 0.680 & 0.874 & 0.867 \\
Pure ($r$)                  & 0.062 & 0.052 & 0.525 & 0.527 \\
AUC $\downarrow$ (over $r$) & 0.246 & \textbf{0.220} & 0.769 & \textbf{0.767} \\
\bottomrule
\end{tabular}

\vskip -0.1in
\end{table}

While RQ1 demonstrates consistent downstream improvements from inserting VEPA, such gains could in principle arise from incidental factors, such as additional optimization or strengthened language priors.
RQ3 therefore examines whether VEPA shifts model behavior toward greater reliance on valid visual input, rather than improving performance through language-only shortcuts.

We evaluate visual reliance using two benchmarks.
ChartQA~\citep{masry2022chartqa} represents a strongly vision-dependent setting that requires reading chart-specific visual content, whereas GQA~\citep{hudson2019gqa} serves as a broad VQA benchmark for which prior work reports substantial question-only accuracy, indicating that language priors can partially support answering.
Both models, standard SFT and VEPA followed by SFT, are trained on the same A-OKVQA SFT data to control for the supervised signal.
To probe reliance on visual input, we perform counterfactual evaluations by corrupting images at inference time.
In addition to clean images, we consider three corruption settings: Gaussian blur, partial noise, and pure noise.
Table~\ref{tab:rq3_corruption} reports clean accuracy and retention under each corruption type, where retention is defined as
$r \triangleq \mathrm{Acc}_{\mathrm{corrupt}} / \mathrm{Acc}_{\mathrm{clean}}$.
We further summarize robustness using the AUC computed from $(r_{\text{blur}}, r_{\text{partial}}, r_{\text{pure}})$ in order; lower AUC indicates stronger reliance on valid visual input.
Under clean images, VEPA improves accuracy on both benchmarks.

On ChartQA, image corruption leads to substantial performance degradation, and VEPA exhibits lower retention and lower retention AUC than standard SFT.
Given the strong visual dependency of ChartQA, this pattern indicates that VEPA relies more heavily on visual evidence rather than maintaining performance via language priors when the image becomes unreliable.
The effect is most pronounced under partial noise, where visual information remains partially informative and grounding is still actionable.
On GQA, VEPA closely matches the retention behavior of standard SFT while improving clean accuracy.
This is consistent with GQA serving as a general-purpose benchmark in which language priors are known to contribute to performance.
Taken together, these results indicate that VEPA selectively increases reliance on visual input when vision is essential, without degrading robustness in broader VQA settings.

\obsbox{\textbf{Insight~III.} Counterfactual image corruption reveals that VEPA increases reliance on valid visual input rather than amplifying language priors.}

\subsection{RQ4. Training Dynamics}
\begin{figure}[t]
  \centering

  \includegraphics[width=\columnwidth]{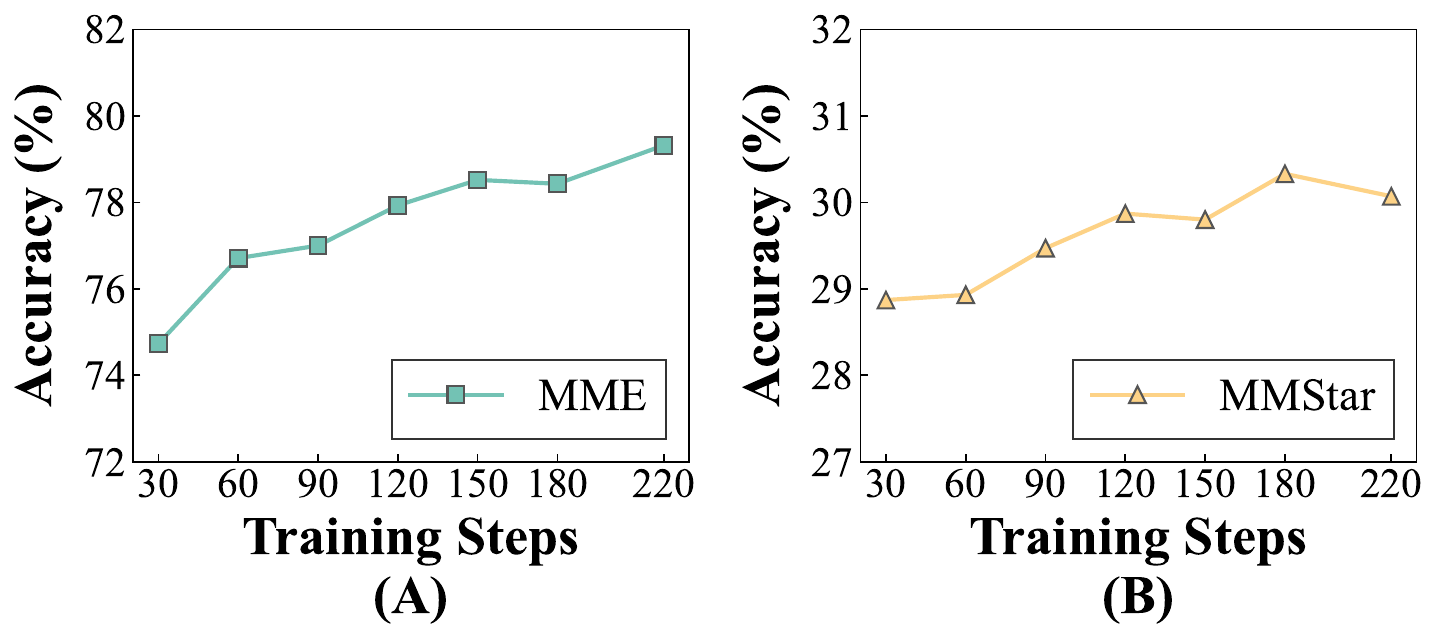}

  \caption{
    \textbf{Performance evolution during VEPA.}
    (A) illustrates the Accuracy trends on the MME dataset, while (B) presents the corresponding results on the MMStar dataset.
  }
  \label{fig:checkpoint-evolution}
  \vspace{-1em}
\end{figure}
To analyze how visual grounding evolves during VEPA, we periodically evaluate intermediate policy checkpoints throughout training.
At each checkpoint, we measure zero-shot accuracy on MME and MMStar, using identical evaluation and decoding settings.

Figure~\ref{fig:checkpoint-evolution} shows a clear upward performance trend as training progresses.
Accuracy on MME increases from 74.73\% to 79.28\%, while MMStar improves from 21.53\% to 24.87\%, with only minor non-monotonic fluctuations.
The smooth improvement trajectory indicates that VEPA produces stable, incremental gains rather than relying on abrupt phase transitions.
Notably, the rate of improvement is highest during the early stage of training and gradually saturates.
This pattern suggests that VEPA first rapidly improves coarse perceptual alignment, after which subsequent updates primarily refine evidence selectivity and training stability.

Overall, these dynamics support the view that VEPA progressively strengthens visual grounding throughout training, consistent with the evidence-sufficiency objective and the blind-reader reward design.

\obsbox{\textbf{Insight~IV.} Visual grounding improves progressively during VEPA training, with early rapid gains followed by later-stage refinement, indicating stable and cumulative perceptual alignment.}

\vspace{-0.5em}
\subsection{Sensitivity Analysis}
\vspace{-0.5em}

We analyze the sensitivity of VEPA along two dimensions: (i) auxiliary blind-reader capacity and (ii) reinforcement learning data scale.

\noindent\textbf{Blind-reader capacity.}
To evaluate dependence on the blind-reader strength, we replace the default Qwen2.5-7B-Instruct auxiliary model with a smaller Qwen2.5-3B-Instruct model while keeping all other VEPA settings unchanged.
Table~\ref{tab:aux_ablation} shows that VEPA remains effective with the reduced auxiliary capacity.
Without downstream SFT, VEPA improves MMStar accuracy from 28.60\% to 32.33\% and GQA from 45.07\% to 53.40\%.
When followed by A-OKVQA SFT, both auxiliary choices yield comparable final performance, reaching 43.00\% on MMStar and 54.57\% on GQA with the 3B auxiliary model, versus 42.80\% and 54.75\% with the 7B auxiliary model.
These results indicate that VEPA does not critically depend on a large auxiliary model and remains robust across auxiliary capacity.

\begin{figure}[t]
  \centering

  \includegraphics[width=\columnwidth]{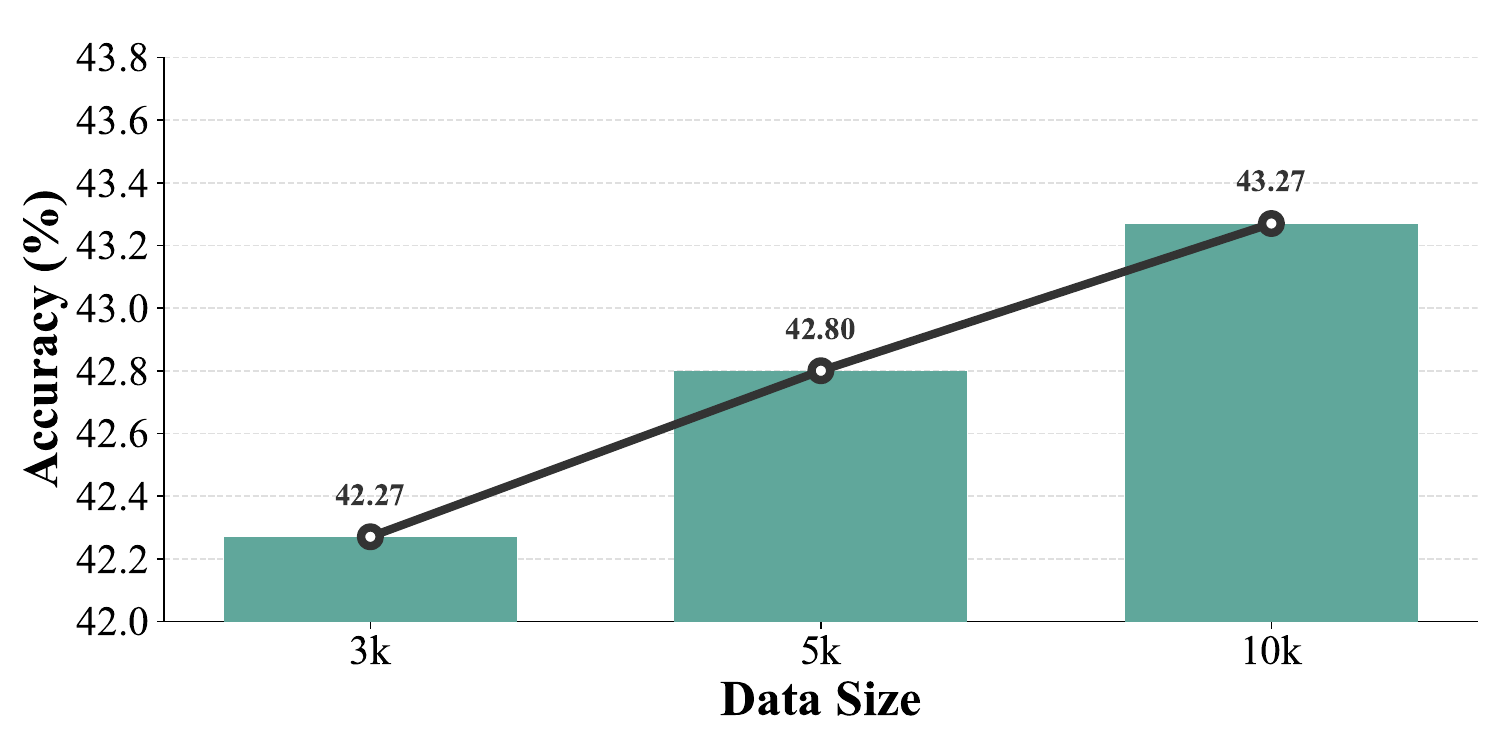}

  \caption{
    \textbf{Data size ablation.} We scale training data from 3k to 10k samples.The results demonstrate a consistent improvement in accuracy as the data size increases.
  }
  \label{fig:rldataset}
  \vspace{-1em}
\end{figure}
\textbf{RL data scale.}
As illustrated in~\ref{fig:rldataset} we further study data efficiency by constructing VEPA RL training sets of size $N \in \{3\text{k}, 5\text{k}, 10\text{k}\}$ via subsampling from the same filtered data mixture.
VEPA remains effective even with a lightweight RL dataset (3k examples), while performance continues to improve as the dataset scales to 10k.
This suggests that VEPA is both data-efficient and scalable, with additional headroom available under larger RL data regimes.

\begin{table}[!htbp]
\caption{\textbf{Auxiliary model sensitivity.} Accuracy (\%) on MMStar and GQA for VEPA trained with various blind readers.}
\label{tab:aux_ablation}
\centering
\small
\setlength{\tabcolsep}{5pt}
\renewcommand{\arraystretch}{1.08}

\begin{tabular}{l cc cc}
\toprule
\multirow{2}{*}{\textbf{Dataset}} &
\multicolumn{2}{c}{\textbf{3B aux}} &
\multicolumn{2}{c}{\textbf{7B aux}} \\
\cmidrule(lr){2-3}\cmidrule(lr){4-5}
& \textbf{VEPA} & \textbf{VEPA+SFT} & \textbf{VEPA} & \textbf{VEPA+SFT} \\
\midrule
MMStar & 32.33 & \textbf{43.00} & 30.07 & \textbf{42.80} \\
GQA    & 53.40 & \textbf{54.57} & 51.46 & \textbf{54.75} \\
\bottomrule
\end{tabular}

\vskip -0.1in
\end{table}

\vspace{-0.5em}
\section{Conclusion}
\vspace{-0.5em}

We introduced \textbf{Visual Evidence Pre-Alignment (VEPA)}, an intermediate training stage inserted between caption-based pretraining and downstream post-training to explicitly strengthen visual grounding before task-level reasoning.
Consistent with the principle of \emph{see first, answer later}, VEPA encourages models to extract question-conditioned visual evidence prior to answer generation, improving how visual information is utilized during inference.
Methodologically, VEPA introduces a sufficiency-driven reinforcement learning objective that enables perceptual alignment using standard VQA supervision, without requiring additional evidence annotations.
Across diverse benchmarks and training settings, VEPA consistently improves performance on visually demanding tasks while preserving in-domain accuracy.
Further analyses indicate that these improvements are associated with more reliable and transferable visual grounding.
Overall, these findings suggest that explicitly separating perceptual alignment from downstream reasoning provides a practical and scalable direction for improving multimodal models.
This perspective suggests a promising direction for future multimodal training paradigms that explicitly structure perceptual alignment as a first-class objective alongside reasoning optimization.

\section*{Impact Statement}
This paper presents work whose goal is to advance the field of Machine
Learning. There are many potential societal consequences of our work, none
which we feel must be specifically highlighted here.

\bibliographystyle{icml2026}
\bibliography{main}

\newpage
\appendix
\onecolumn
\section{Appendix}

\subsection{Experiment Settings}

\label{appdix:experiment details}

\paragraph{Dataset and Benchmarks.} To strictly evaluate the performance of our proposed VEPA, we conduct experiments on comprehensive benchmarks that assess diverse multimodal capabilities.
We first include GQA~\cite{hudson2019gqa} and A-OKVQA~\cite{schwenk2022okvqa} to evaluate compositional visual reasoning and knowledge-intensive question answering, respectively.
Although GQA applies bias-mitigation via answer-distribution smoothing, it still retains non-trivial language priors: Hudson \yrcite{hudson2019gqa} report that a question-only LSTM baseline achieves 42.1\% accuracy, indicating that a non-negligible portion of questions can be partially resolved from the question text alone.

To assess the model's ability to process fine-grained visual information, we utilize TextVQA~\cite{singh2019towards} for optical character recognition in natural scenes and ChartQA~\cite{masry2022chartqa} for logical reasoning over complex charts.
To further extend the evaluation to scientific and geometric domains, we incorporate ScienceQA~\cite{lu2022learn}, AI2D-Merge~\cite{kembhavi2016diagram}, and Geo3K~\cite{lu2021inter}, which challenge the model with textbook-grade diagrams and mathematical reasoning.
Additionally, we integrate CLEVR~\cite{johnson2017clevr} and FineVision~\cite{wiedmann2025finevision} to strictly test synthetic compositional logic and fine-grained visual discrimination, respectively.
For a holistic evaluation of MLLMs, we adopt MME~\cite{fu2025mme}, which covers a broad range of perception and cognition tasks, and MMStar~\cite{chen2024we}, a benchmark specifically curated to test models on hard samples across multiple disciplines.
Finally, we employ POPE~\cite{li2023evaluating} to specifically measure the object hallucination rates and the robustness of the generated responses.

\paragraph{Models.} In our experiments, we primarily utilize Qwen2-VL~\cite{wang2024qwen2} as the backbone architecture for training our VEPA framework.
This model is selected for its state-of-the-art performance in visual understanding and its capability to handle arbitrary image resolutions through dynamic resolution support.
To assist with auxiliary tasks such as data processing and response refinement, we incorporate a suite of lightweight yet capable language models.
Specifically, we employ the 3B and 7B variants of the Qwen2.5 series~\cite{Yang2024Qwen25TR}, which offer a strong balance between efficiency and reasoning capability.
Furthermore, we adopt GPT-5-nano~\cite{openai_gpt5_2025}, the most efficient variant in the GPT-5 family, which is specifically optimized for high-throughput instruction following and low-latency applications, serving as a robust baseline for commercial lightweight systems.

\paragraph{Evaluation Metrics.} We report Accuracy as the primary metric across all benchmarks.
For open-ended generation tasks, we implement a deterministic matching protocol to address linguistic variations.
Specifically, both predictions and ground truths undergo normalization, including case lowering, punctuation removal, and stop-word stripping.
To further handle morphological discrepancies, we expand the ground truth into its inflectional variants (e.g., singular and plural forms) to verify semantic equivalence against the prediction.
For multiple-choice tasks, we employ a hierarchical parsing strategy that prioritizes the extraction of explicit option labels.
When labels are absent, the evaluation falls back to semantic content matching, which verifies the presence of the correct option's text.
Crucially, this content matching enforces an exclusivity constraint: a prediction is considered correct only if it contains the target content without including text from incorrect distractors, thereby preventing false positives from hallucinated candidates.

\paragraph{Implementation Details.} We implement our proposed VEPA framework using Group Relative Policy Optimization (GRPO) to fine-tune the Qwen2-VL-2B~\cite{wang2024qwen2} backbone.
The training process leverages Fully Sharded Data Parallel (FSDP) with bfloat16 precision to maximize computational efficiency.
We employ the AdamW optimizer with a constant learning rate of $1\times 10^{-6}$ following a 5\% warmup phase, and set the KL divergence coefficient $\beta$ to 0.01 to maintain policy stability.
For data generation, we utilize the VLLM engine to sample $G=4$ candidate captions for each visual query with a temperature of 0.9.
Crucially, our reward mechanism incorporates a lightweight auxiliary judge (Qwen2.5-7B-Instruct) that evaluates responses based on three dimensions: correctness, honesty, and fluency.
Specifically, we assign a positive reward (+1.0) for factually correct answers derived from visual evidence, while imposing strict penalties for "cheating" behaviors—such as outputting the answer directly without description (-1.0)—and for linguistic repetition (-0.5).
The system prompt is carefully engineered to simulate a ``blind reader'' scenario, compelling the model to function as an objective visual analyst rather than a direct question answerer.

\subsection{Prompts Used in Experiments}
\label{subsec:prompts}
Below are the prompts used in the experiments.

\textbf{Caption Generation} elicits \emph{question-conditioned visual evidence}: the model is instructed to produce a comprehensive, objective description of visible elements (including text and spatial relations) that are relevant to the question, while \emph{explicitly forbidding} direct answers, so that the output can serve as standalone evidence for a blind solver.
\textbf{Auxiliary Judge} instantiates our blind-reader mechanism: given the generated evidence and the question, it first performs \emph{cheat detection} (flagging outputs that answer without visual grounding), and then attempts to \emph{solve} the question using only the provided description; its structured output is used to reward grounded evidence and penalize answer leakage.
Finally, \textbf{VQA Judge} is used for automatic evaluation: it compares the model prediction against the ground-truth answer under semantic-equivalence rules and returns a single Boolean decision (\texttt{TRUE}/\texttt{FALSE}), enabling scalable and reproducible scoring.

\begin{tcolorbox}[sharp corners, breakable, colframe=teal, colback=white,
       boxrule=3pt, boxsep=0.5pt, enhanced,
       shadow={3pt}{-3pt}{0pt}{opacity=1,mygrey},
       title={Caption Generation}]
       \label{1}
\footnotesize
{
You are an expert visual analyst. Given an image and a question, your goal is to provide a comprehensive and detailed objective description of the visual content.

\textbf{Context:} This description will be used by a blind person to reason about and answer the question. Therefore, your description must act as the raw visual evidence required for reasoning.

\textbf{Strict Guidelines:}

1. Focus on Visual Evidence, NOT Conclusions: Describe the visible elements, text, spatial relationships, colors, and specific details that are relevant to the question. Do not just state the final answer.

2. Be Comprehensive: Do not be concise. Include ALL visual details, no matter how small, if they are necessary to support the reasoning process for the question.

3. No Direct Answers: You must NOT explicitly output the answer(e.g., do not say ``The answer is B", ``It is a cat'' or ``Select Option A"). Instead, provide all the necessary visual evidence so that a blind person can deduce the answer themselves.

4. Format: Output ONLY the detailed caption.

}
\end{tcolorbox}

\begin{tcolorbox}[sharp corners, breakable, colframe=teal, colback=white,
       boxrule=3pt, boxsep=0.5pt, enhanced,
       shadow={3pt}{-3pt}{0pt}{opacity=1,mygrey},
       title={Auxiliary Judge}]
       \label{2}
\footnotesize
{

You are an impartial judge and a blind solver. I will provide a `Visual
Description' generated by an AI and a `Question'.

\textbf{Task 1: Cheat Detection}

Check if the description is trying to cheat. `Cheating' means:

- Directly explicitly stating `The answer is X' WITHOUT visual context.

- Providing the answer directly (e.g., just saying `Yes', `2', or `Blue')
  WITHOUT any descriptive sentence.

Important Exception: It is NOT cheating if the description states a visual
fact that happens to be the answer.

\textbf{Task 2: Solve}

Use the description to answer the question briefly.

\textbf{Output Format (CRITICAL):}

If you detect cheating, start with: ``[[CHEAT]]'' and then the answer.

If not, start with: ``[[honest]]'' and give the answer.

Visual Description: $\{caption\}$

Question: $\{question\}$

}
\end{tcolorbox}

\begin{tcolorbox}[sharp corners, breakable, colframe=teal, colback=white,
       boxrule=3pt, boxsep=0.5pt, enhanced,
       shadow={3pt}{-3pt}{0pt}{opacity=1,mygrey},
       title={VQA Judge}]
       \label{3}
\footnotesize
{
You are an expert VQA (Visual Question Answering) judge.

\textbf{Your Task:}

Evaluate if the model's PREDICTION is correct based on the GROUND TRUTH (GT).

\textbf{Judgment Rules (Important):}

1. Core Meaning: Focus on semantic meaning.

2. Ignore Trivialities: Ignore capitalization, punctuation, minor phrasing.

3. Rationale Check: If GT has explanation and Prediction follows similar
   reasoning logic, count as CORRECT.

\textbf{Data:}

- Question: $\{question\}$

- Ground Truth: $\{gt\}$

- Prediction: $\{prediction\}$

\textbf{Output Requirement:}

You must output ONLY one word.

- Output `TRUE' if the prediction is semantically correct.

- Output `FALSE' if the prediction is incorrect.

Do NOT output JSON. Do NOT output any explanation.

}
\end{tcolorbox}

\subsection{More Experiment Results on Evidence Quality}
\begin{figure}[!htbp]
  \centering

  \includegraphics[width=\columnwidth]{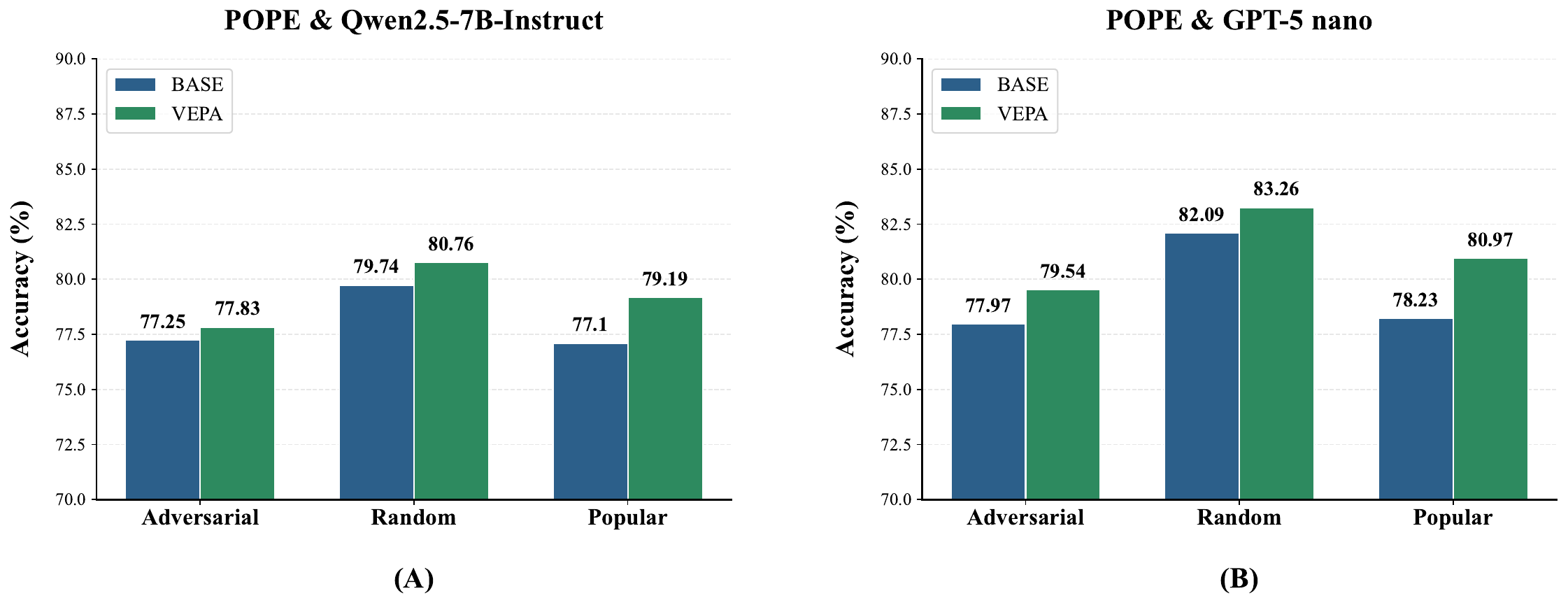}

  \caption{
    Detailed accuracy comparison on the POPE dataset across three categories: Adversarial, Random, and Popular.
    (A) displays results for \textbf{Qwen2.5-7B-Instruct}, while (B) shows \textbf{GPT-5 Nano}.
  }
  \label{fig:pope-detailed-comparison}
  \vspace{-0.1in}
\end{figure}

\label{app:aux_model_variation}

This section reports additional evidence-quality results complementing \S\ref{subsec:rq2}.
Beyond the aggregate scores, we further decompose POPE into its three subsets (Adversarial, Random, Popular) to assess whether VEPA improves evidence sufficiency across query types.

Figure~\ref{fig:pope-detailed-comparison} shows consistent gains in blind-reader accuracy for all POPE subsets.
With \textsc{Qwen2.5-7B-Instruct} as the blind reader, VEPA improves accuracy from 77.25\% to 77.83\% on \textsc{Adversarial}, from 79.74\% to 80.76\% on \textsc{Random}, and from 77.10\% to 79.19\% on \textsc{Popular}.
The largest improvement occurs on \textsc{Popular} (+2.09\%), indicating that VEPA particularly enhances the extraction of salient, question-relevant visual cues, while remaining effective under adversarial instances.

These conclusions are robust to the choice of the blind reader.
Using \textsc{GPT5-Nano} yields the same pattern: 77.97\% to 79.54\% on \textsc{Adversarial}, 82.09\% to 83.26\% on \textsc{Random}, and 78.23\% to 80.97\% on \textsc{Popular}.
The agreement across readers suggests that the improved accuracy reflects stronger evidence sufficiency rather than evaluator-specific artifacts.

Overall, the POPE breakdown reinforces the main finding of RQ2~\ref{subsec:rq2}: VEPA improves the recoverability of task-critical visual cues from generated evidence, thereby strengthening perceptual grounding without relying on dense descriptions.

\subsection{More Experiment results on Performance Analysis}
\label{sec:appendix_mme_mmstar}

To complement the aggregated results presented in Table~\ref{tab:rq1_main}, we provide a detailed performance analysis on the MME and MMStar benchmarks. While the main text reports average scores to demonstrate overall trends, this section breaks down the performance of the \textbf{SFT} and \textbf{VEPA+SFT} settings across specific sub-tasks to evaluate the consistency and robustness of our method.

We specifically analyze ten fine-grained dimensions, comprising seven categories from \textbf{MME} (Color, Count, Numerical Calculation, OCR, Position, Poster, and Text Translation) and four core capability dimensions from \textbf{MMStar} (Coarse Perception, Logical Reasoning, Math, and Science \& Technology).

\begin{figure}[!hbp]
    \centering

    \begin{subfigure}{0.48\textwidth}
        \centering
        \includegraphics[width=\linewidth]{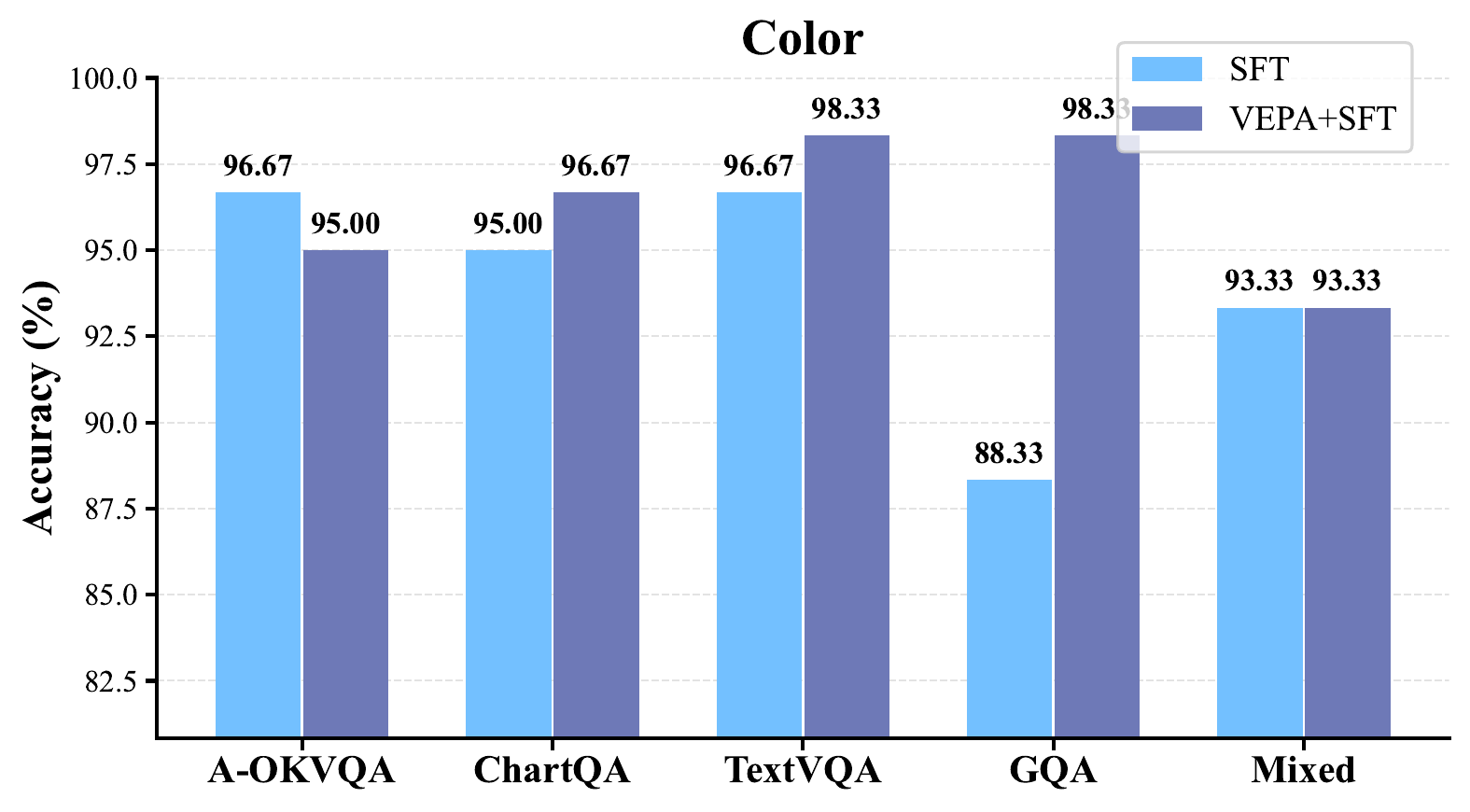}
        \caption{MME: Color}
    \end{subfigure}
    \hfill
    \begin{subfigure}{0.48\textwidth}
        \centering
        \includegraphics[width=\linewidth]{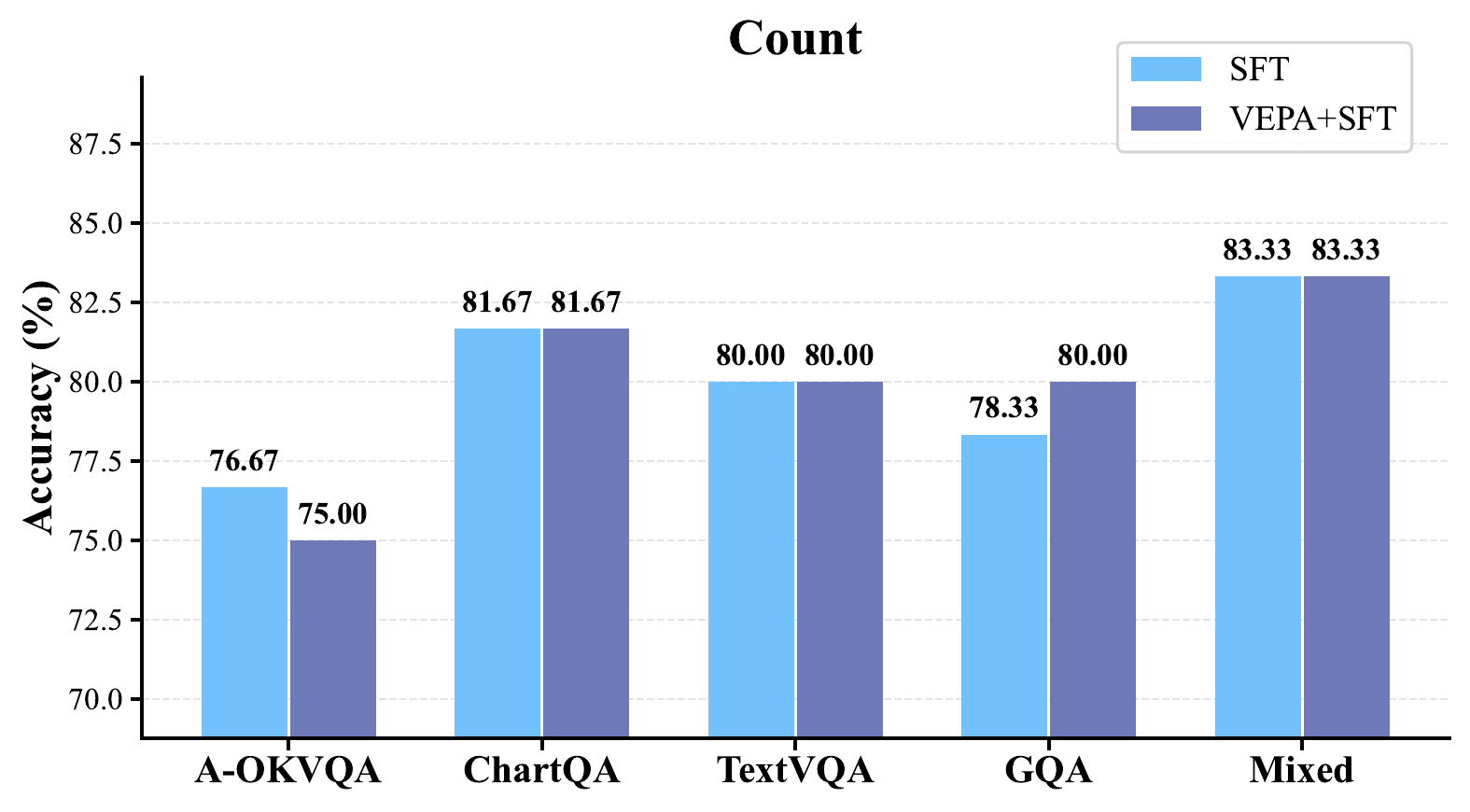}
        \caption{MME: Count}
    \end{subfigure}

    \vspace{1em}

    \begin{subfigure}{0.48\textwidth}
        \centering
        \includegraphics[width=\linewidth]{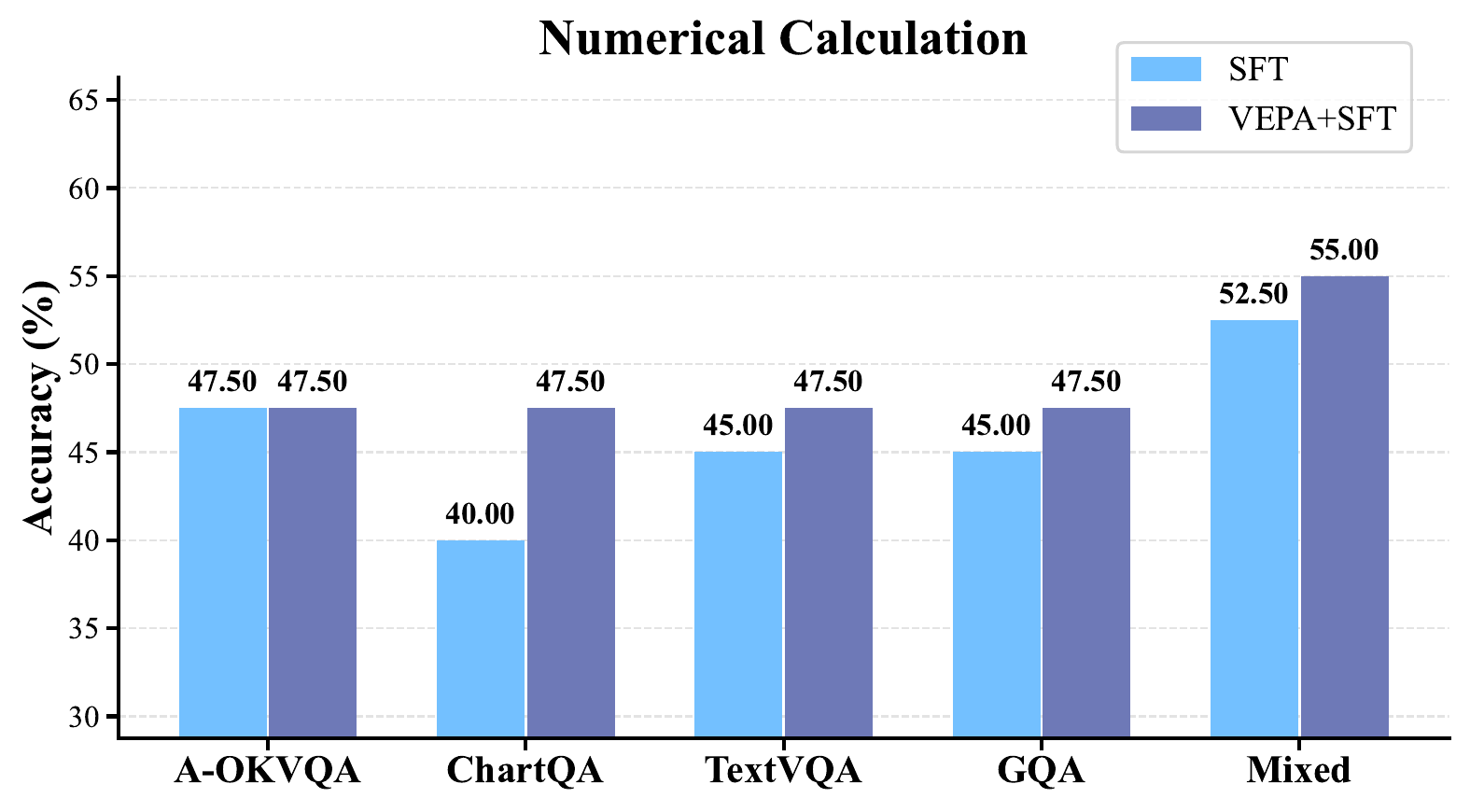}
        \caption{MME: Numerical Calculation}
    \end{subfigure}
    \hfill
    \begin{subfigure}{0.48\textwidth}
        \centering
        \includegraphics[width=\linewidth]{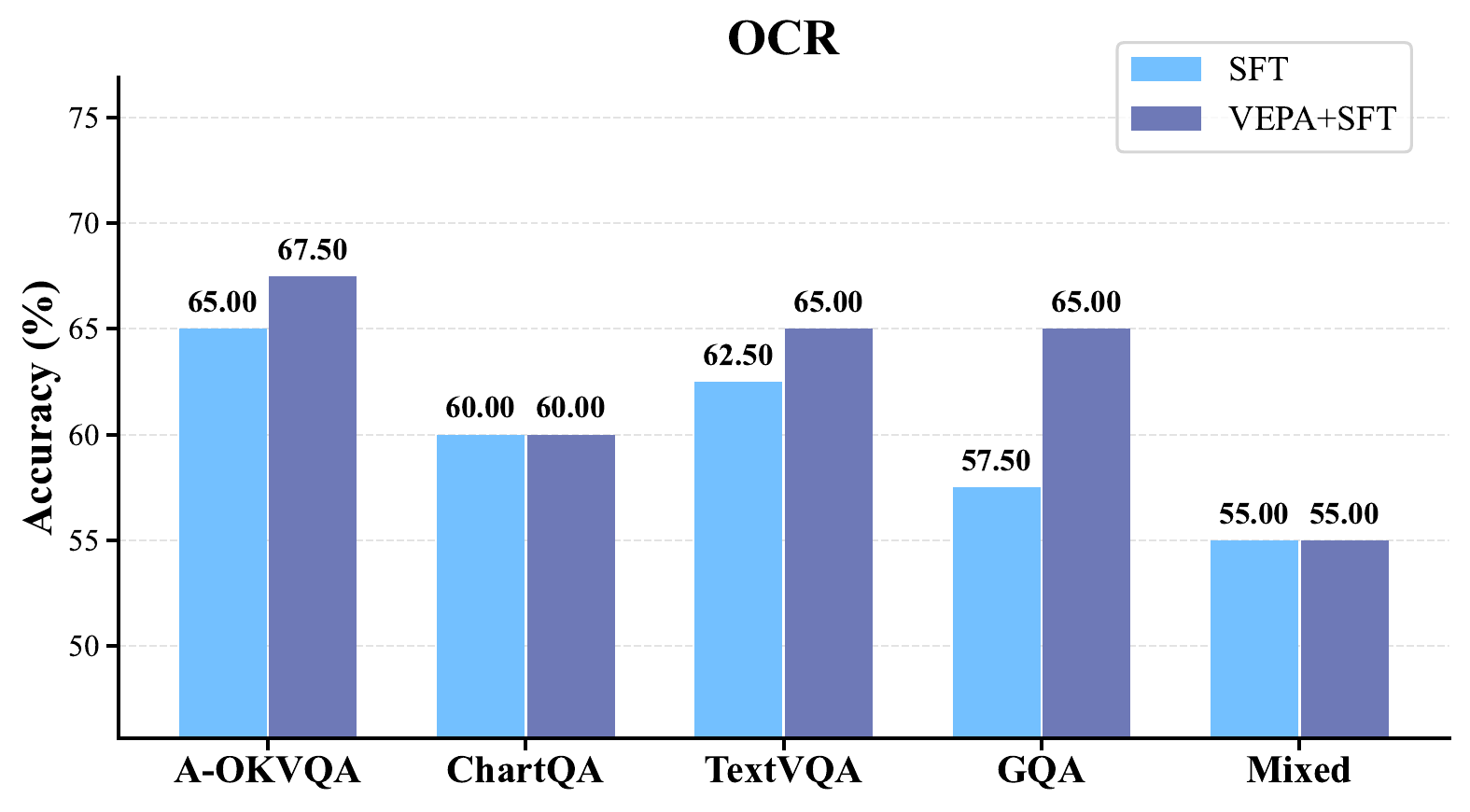}
        \caption{MME: OCR}
    \end{subfigure}

    \vspace{1em}

    \begin{subfigure}{0.48\textwidth}
        \centering
        \includegraphics[width=\linewidth]{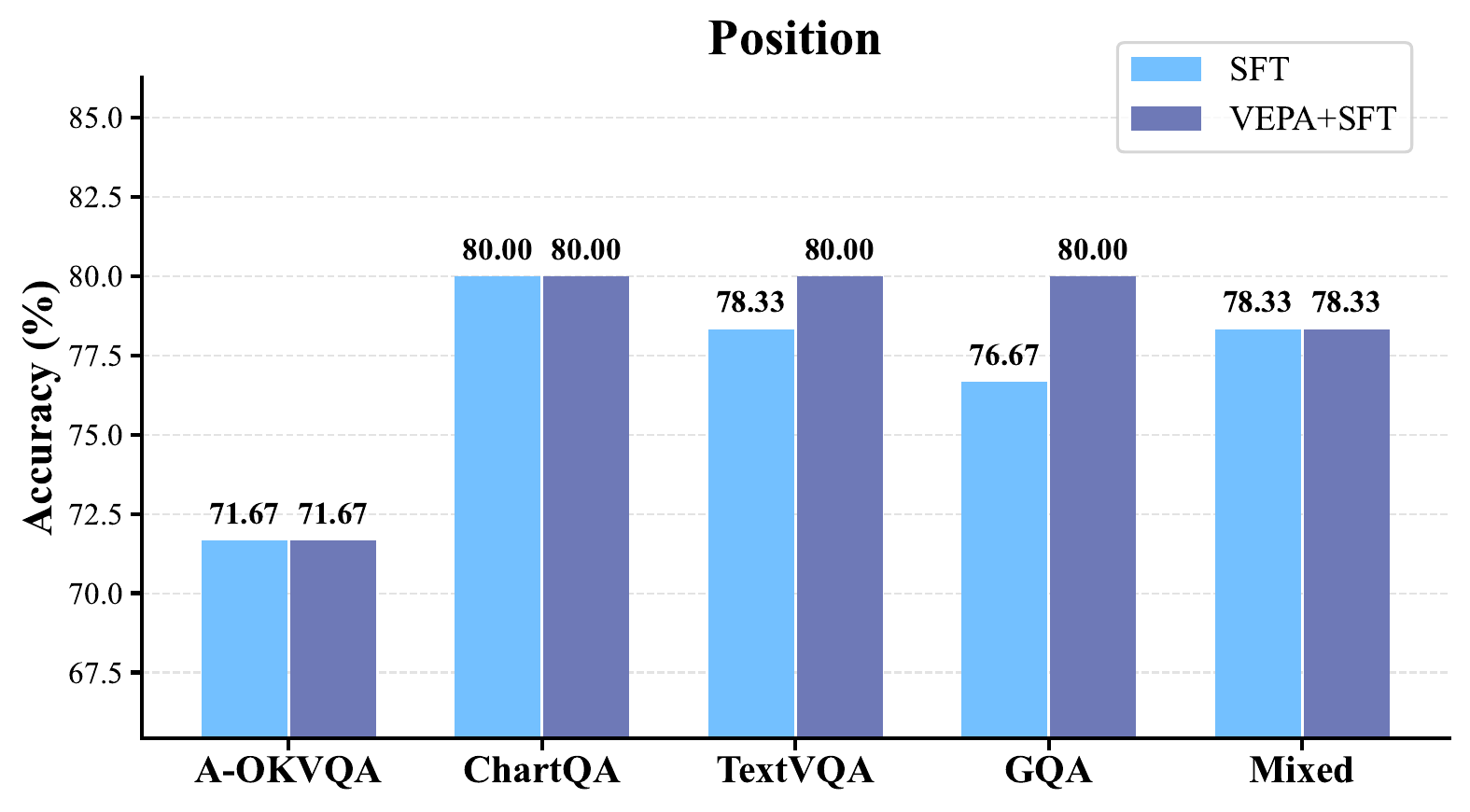}
        \caption{MME: Position}
    \end{subfigure}
    \hfill
    \begin{subfigure}{0.48\textwidth}
        \centering
        \includegraphics[width=\linewidth]{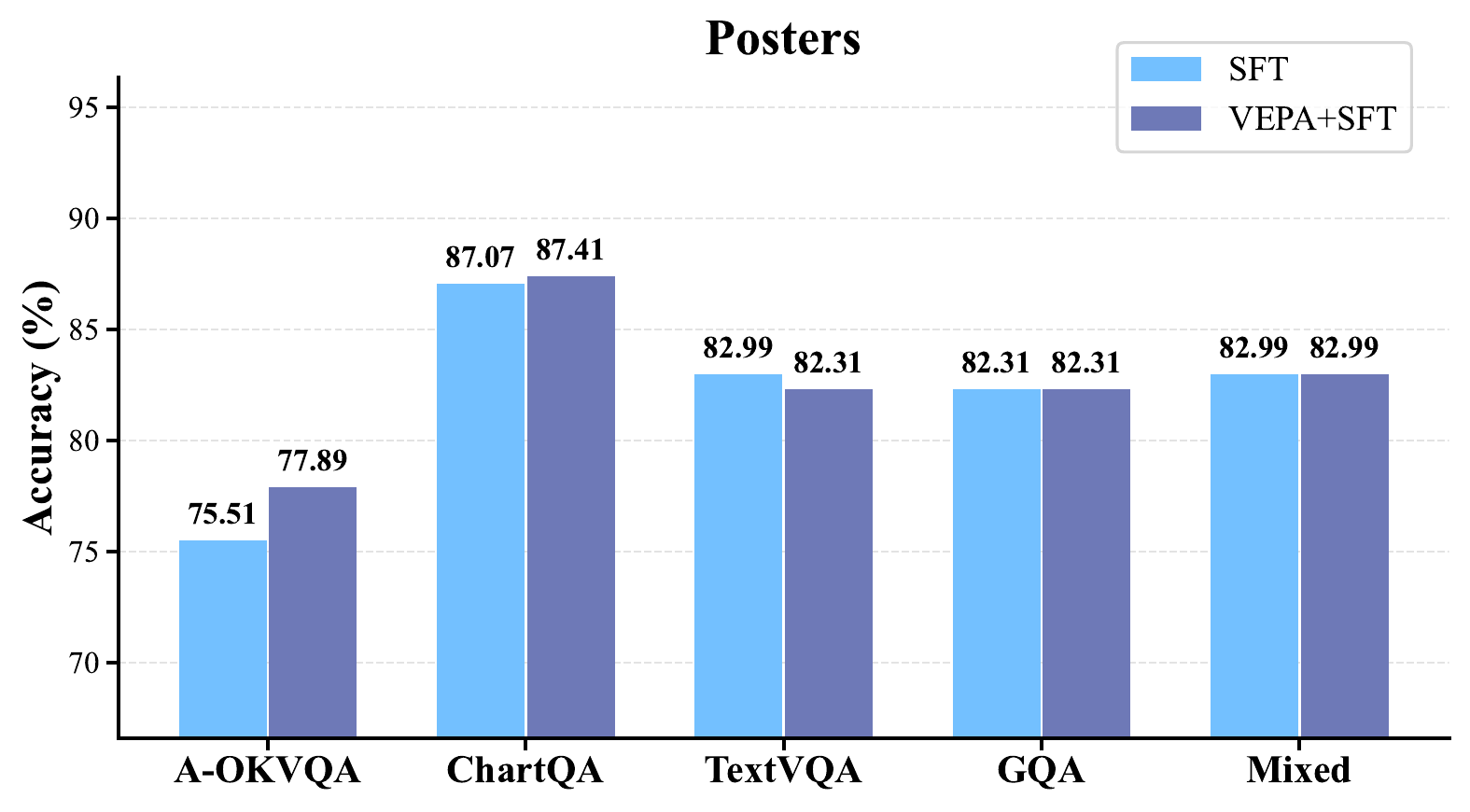}
        \caption{MME: Poster}
    \end{subfigure}

    \vspace{1em}

    \begin{subfigure}{0.48\textwidth}
        \centering
        \includegraphics[width=\linewidth]{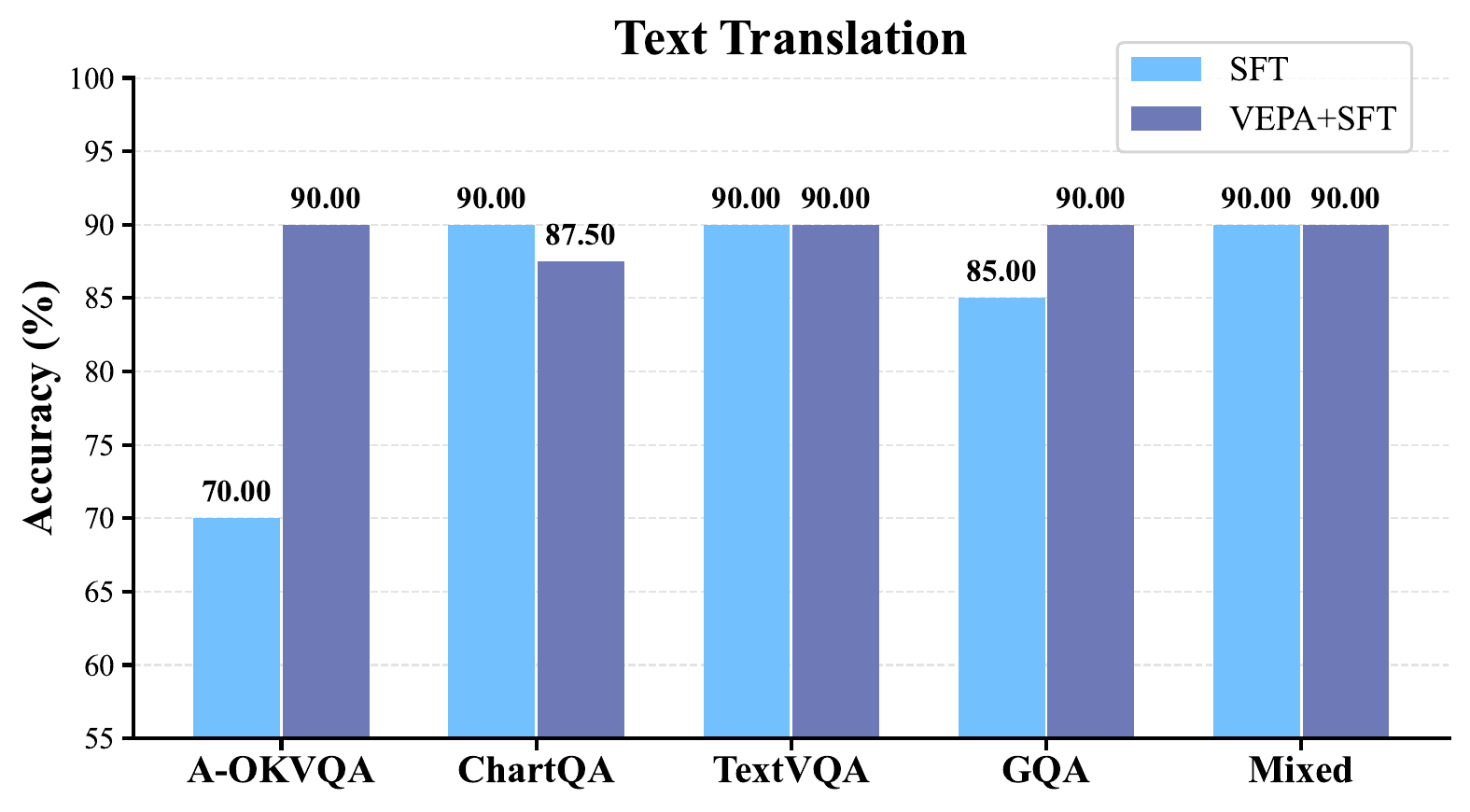}
        \caption{MME: Text Translation}
    \end{subfigure}
    \caption{\textbf{Fine-grained performance breakdown on MME.} We compare the baseline SFT against VEPA+SFT across various sub-tasks. The results demonstrate that VEPA provides consistent improvements across most categories, particularly in tasks requiring precise visual grounding.}
    \label{fig:mme_breakdown}
\end{figure}

\begin{figure}[!htbp]
    \centering

    \begin{subfigure}{0.48\textwidth}
        \centering
        \includegraphics[width=\linewidth]{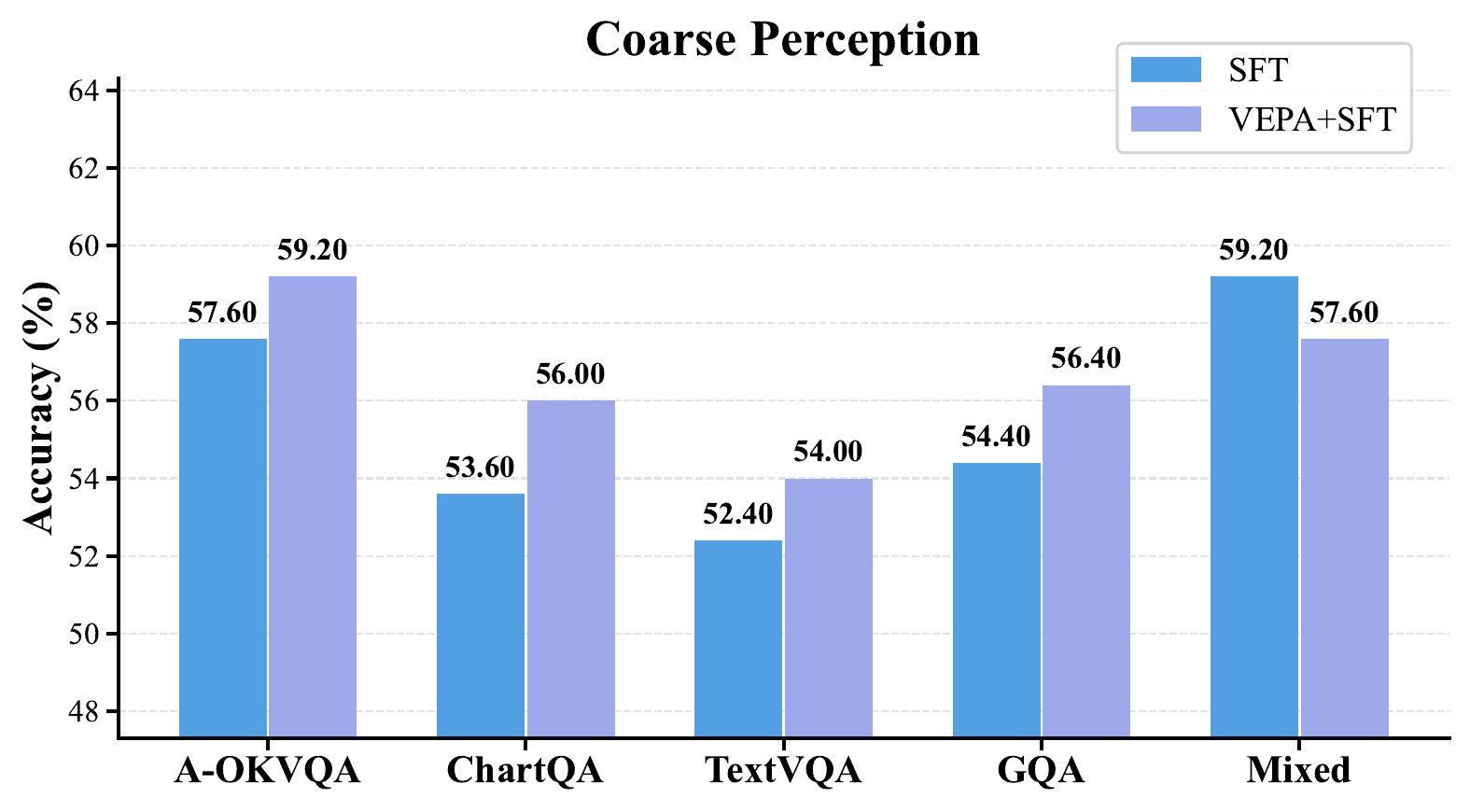}
        \caption{MMStar: Coarse Perception}
    \end{subfigure}
    \hfill
    \begin{subfigure}{0.48\textwidth}
        \centering
        \includegraphics[width=\linewidth]{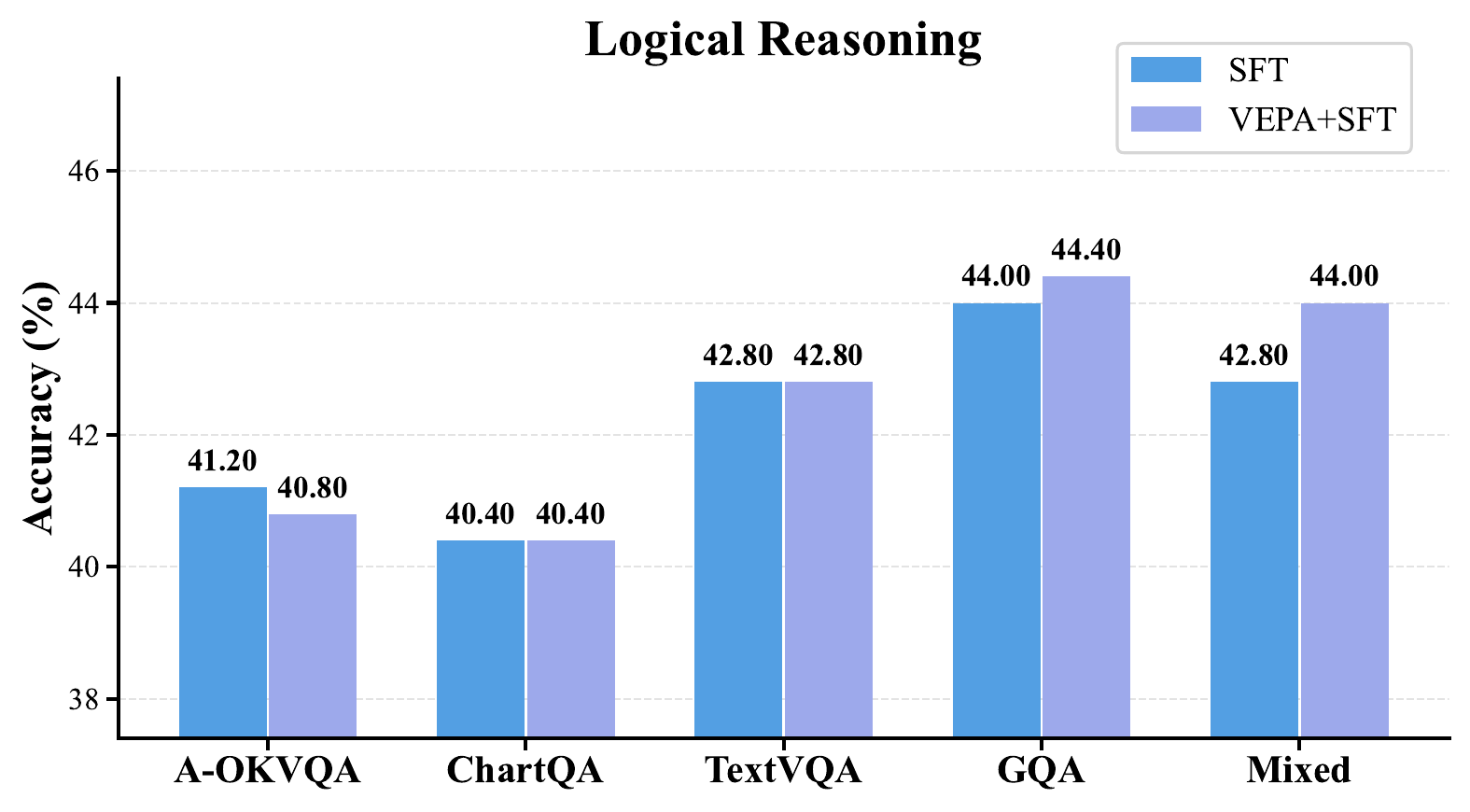}
        \caption{MMStar: Logical Reasoning}
    \end{subfigure}

    \vspace{1em}

    \begin{subfigure}{0.48\textwidth}
        \centering
        \includegraphics[width=\linewidth]{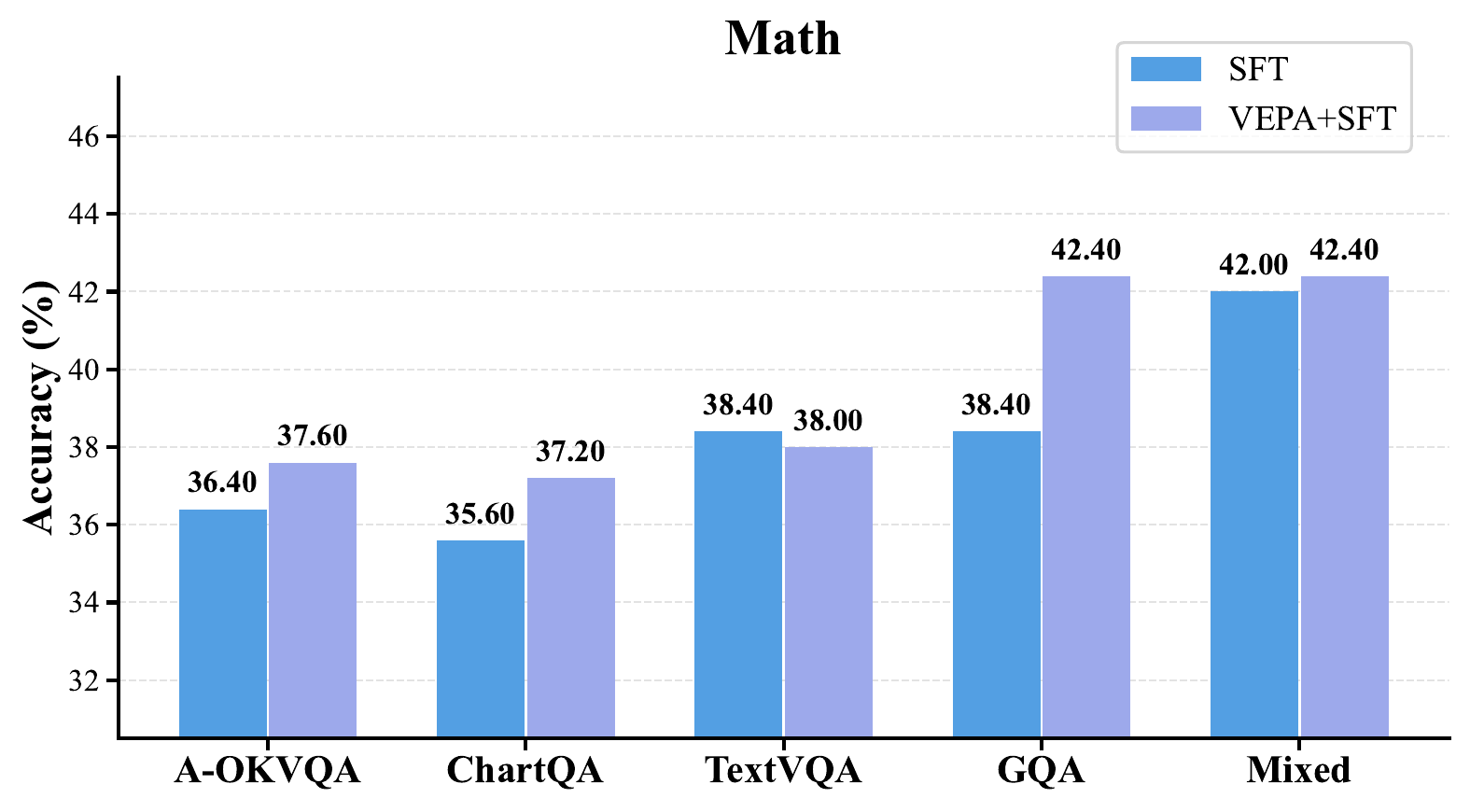}
        \caption{MMStar: Math}
    \end{subfigure}
    \hfill
    \begin{subfigure}{0.48\textwidth}
        \centering
        \includegraphics[width=\linewidth]{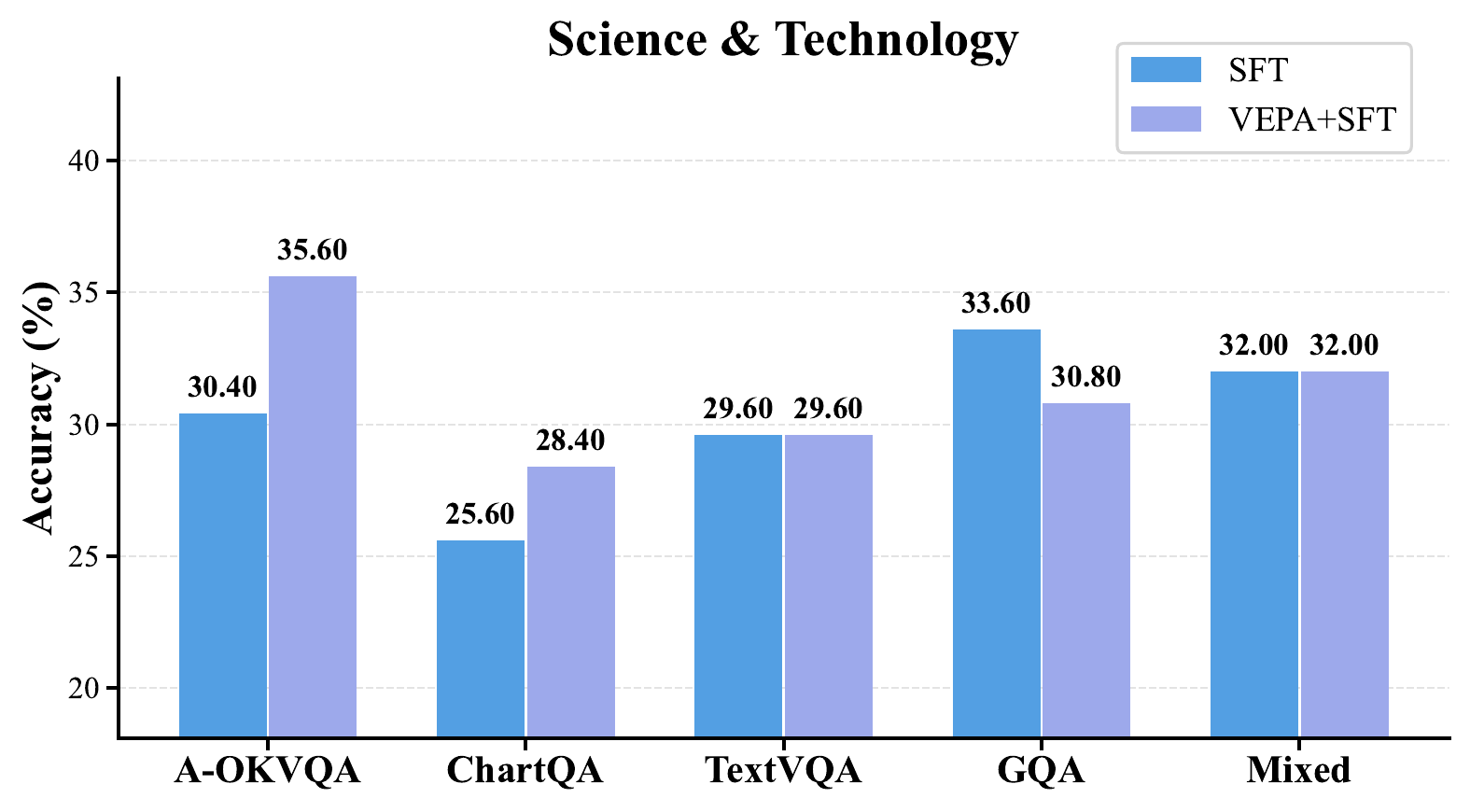}
        \caption{MMStar: Sci-Tech}
    \end{subfigure}

    \caption{\textbf{Fine-grained performance breakdown on MMStar.} We compare the baseline SFT (blue) against VEPA+SFT (green) across various sub-tasks. The results demonstrate that VEPA provides consistent improvements across most categories, particularly in tasks requiring precise visual grounding.}
    \label{fig:mmstar_breakdown}
\end{figure}

Figure~\ref{fig:mme_breakdown} and  Figure~\ref{fig:mmstar_breakdown} visualizes the accuracy comparisons across five distinct SFT data settings (A-OKVQA, ChartQA, TextVQA, GQA, and Mixed). The results reveal the following key observations:

\begin{itemize}
    \item \textbf{Consistent Gains:} Consistent with the averaged results, VEPA+SFT outperforms the standard SFT baseline in the majority of fine-grained categories. Notably, significant improvements are observed in tasks requiring strong visual dependency, such as \textit{Numerical Calculation}, \textit{OCR}, and \textit{Coarse Perception}. This supports our hypothesis that the VEPA stage enhances the model's ability to ground textual generation in visual evidence.

    \item \textbf{Robustness:} While we observe minor performance fluctuations in a few specific settings (e.g., slight regressions in select isolated metrics), the overall performance landscape remains stable. VEPA demonstrates robustness across diverse tasks, indicating that the method improves general multimodal capabilities without trading off performance in specific domains.
\end{itemize}

These fine-grained statistics further corroborate that VEPA serves as an effective intermediate stage, fostering transferable perceptual skills that generalize beyond the specific distribution of the SFT data.

\subsection{More Experiment Results on Sensitivity of Dataset}
\label{subsec:sensitivity of dataset}

We vary the number of RL training instances used in VEPA by sampling 3k, 5k, or 10k examples from the same filtered mixture.
Figure~\ref{fig:rldataset} illustrates accuracy after VEPA followed by A-OKVQA-based SFT.
Performance on MMStar increases monotonically from 42.27\% to 42.80\% and 43.27\% as the VEPA data scale grows, suggesting additional headroom from further scaling perception-alignment data.
And below is the detail performance comparison on MMStar sub-categories.

\begin{table*}[!htbp]
  \centering
  \caption{\textbf{Detailed performance comparison on MMStar sub-categories.} We report the accuracy (\%) across varying training scale (3k, 5k, 10k) for both \textbf{VEPA} and \textbf{VEPA + SFT} settings.}
  \label{tab:mmstar-breakdown}
  \begin{tabular}{lcccccc}
    \toprule
    \rowcolor{white}
    \multirow{2}{*}{\textbf{Category}} & \multicolumn{3}{c}{\textbf{VEPA}} & \multicolumn{3}{c}{\textbf{VEPA + SFT}} \\
    \cmidrule(lr){2-4} \cmidrule(lr){5-7}
    \rowcolor{white}
    & \textbf{3k} & \textbf{5k} & \textbf{10k} & \textbf{3k} & \textbf{5k} & \textbf{10k} \\
    \midrule

    Coarse Perception        & 41.60 & 44.80 & 46.40 & 60.80 & 59.20 & 57.60 \\
    \rowcolor{gray!12}
    Fine-grained Perception  & 32.40 & 34.00 & 31.60 & 38.40 & 36.80 & 37.60 \\
    Instance Reasoning       & 38.80 & 38.80 & 39.20 & 43.60 & 47.20 & 46.00 \\
    \rowcolor{gray!12}
    Logical Reasoning        & 28.80 & 25.60 & 26.80 & 41.20 & 40.80 & 42.00 \\
    Math                     & 20.00 & 20.00 & 22.00 & 34.80 & 37.60 & 39.20 \\
    \rowcolor{gray!12}
    Science \& Technology    & 17.20 & 17.20 & 16.80 & 34.80 & 35.20 & 37.20 \\
    \bottomrule
  \end{tabular}
\end{table*}

\subsection{Case Study}
\label{subsec:case_study}

\begin{figure}[!htbp]
  \centering
  \begin{minipage}[t]{0.49\linewidth}
    \centering
    \includegraphics[width=\linewidth]{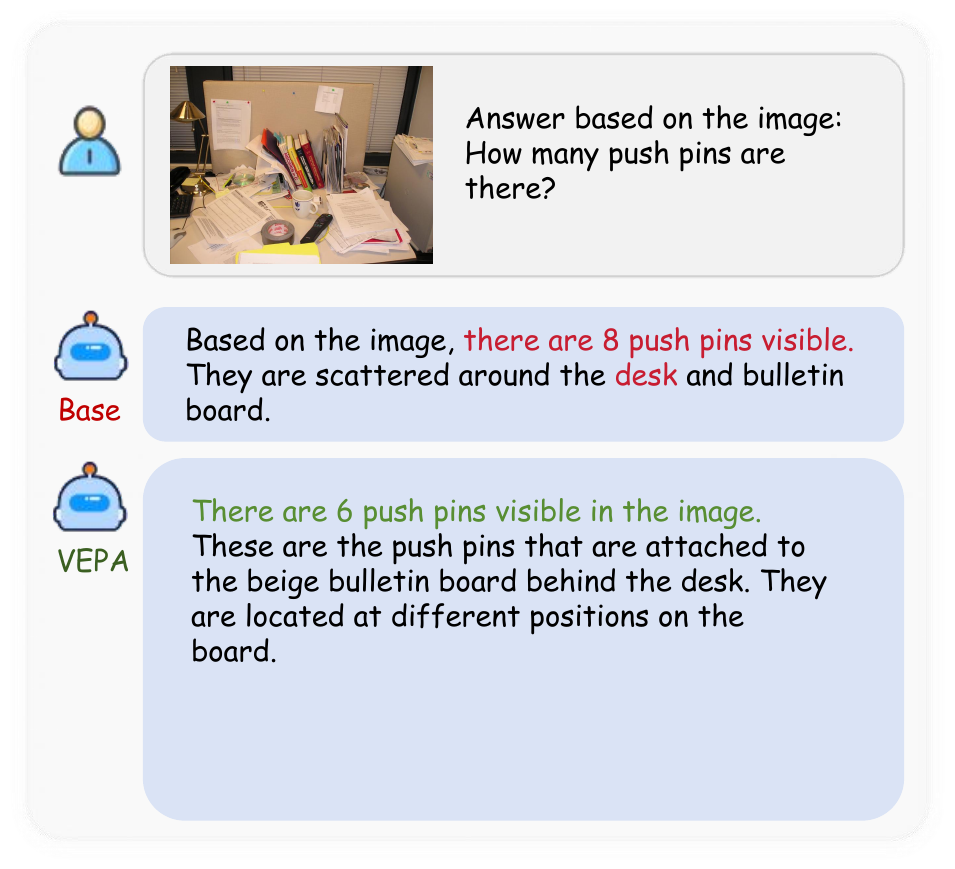}
    \captionof{figure}{Case 1 on Counting}
    \label{fig:case1-counting}
  \end{minipage}
  \hfill
  \begin{minipage}[t]{0.49\linewidth}
    \centering
    \includegraphics[width=\linewidth]{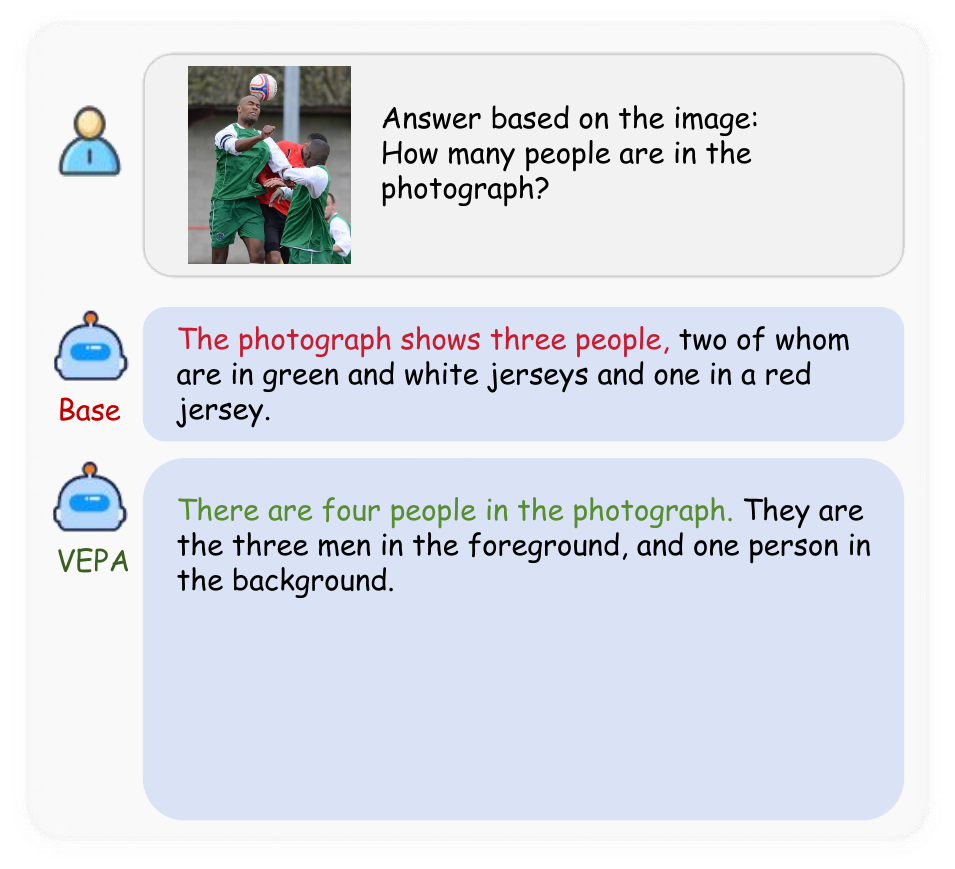}
    \captionof{figure}{Case 2 on Counting}
    \label{fig:case2-counting}
  \end{minipage}
\end{figure}

\begin{figure}[!htbp]
  \centering
  \begin{minipage}[t]{0.49\linewidth}
    \centering
    \includegraphics[width=\linewidth]{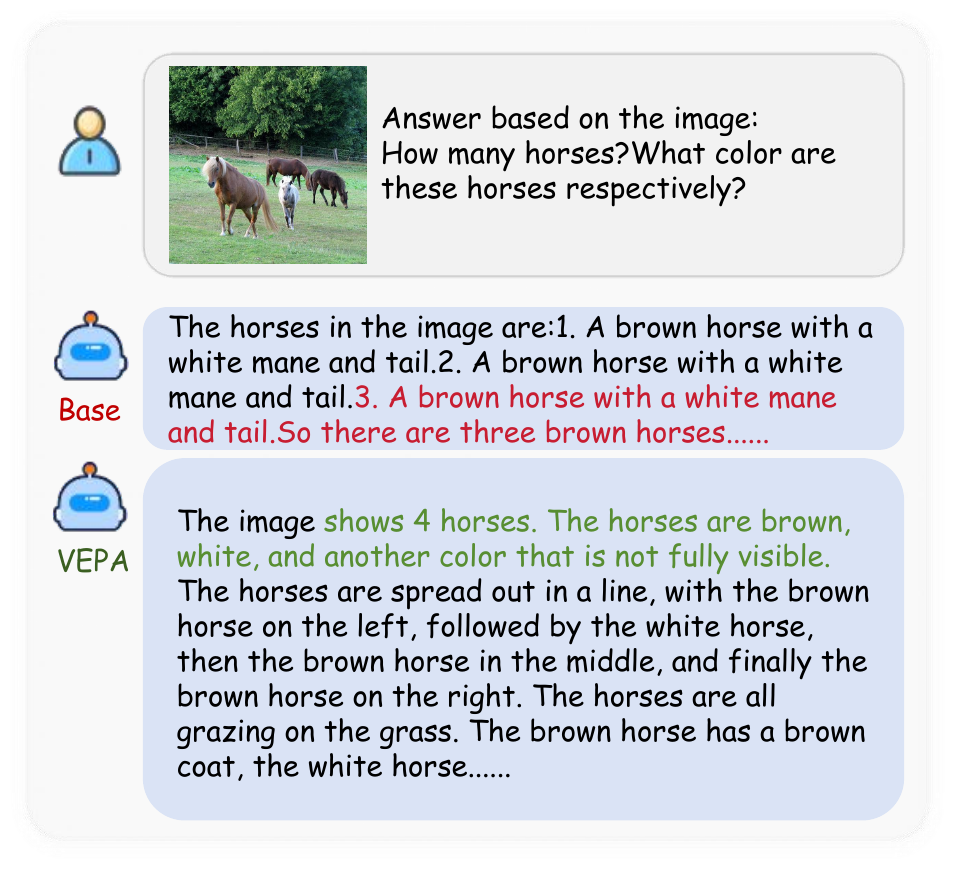}
    \captionof{figure}{Case 3 on Counting and Attribute}
    \label{fig:case3-counting-attribute}
  \end{minipage}
  \hfill
  \begin{minipage}[t]{0.49\linewidth}
    \centering
    \includegraphics[width=\linewidth]{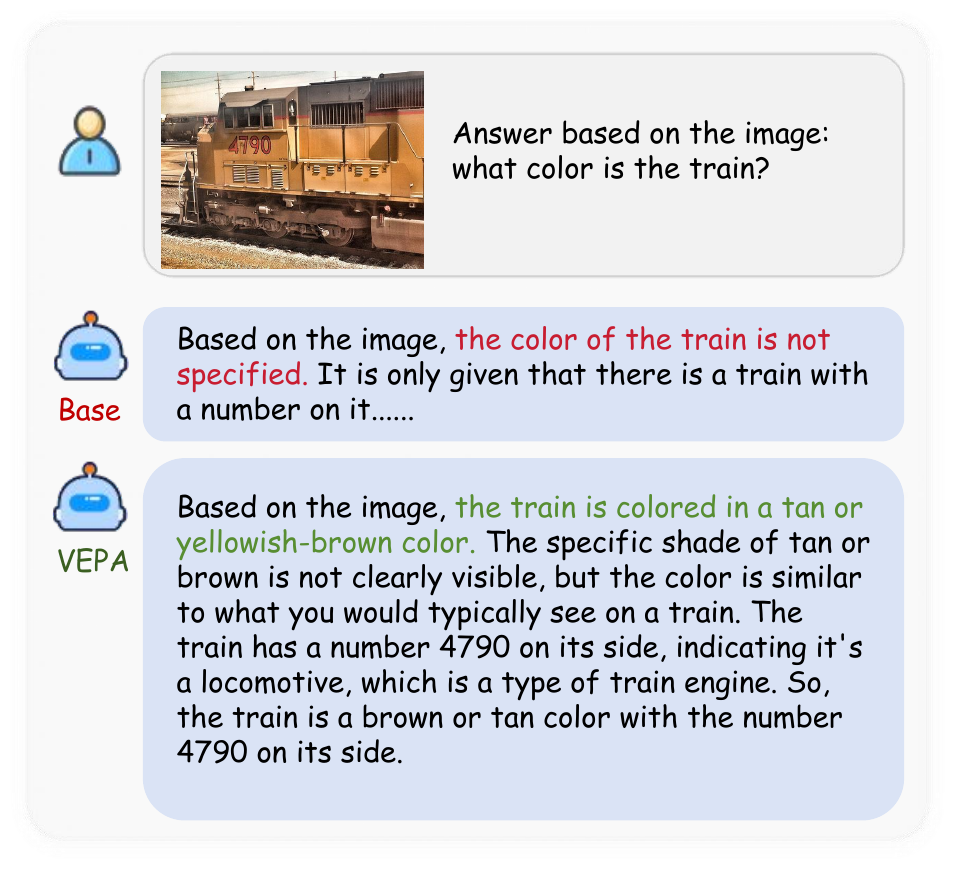}
    \captionof{figure}{Case 4 on Attribute}
    \label{fig:case4-attribute}
  \end{minipage}
\end{figure}

\begin{figure}[!htbp]
  \centering
  \begin{minipage}[t]{0.49\linewidth}
    \centering
    \includegraphics[width=\linewidth]{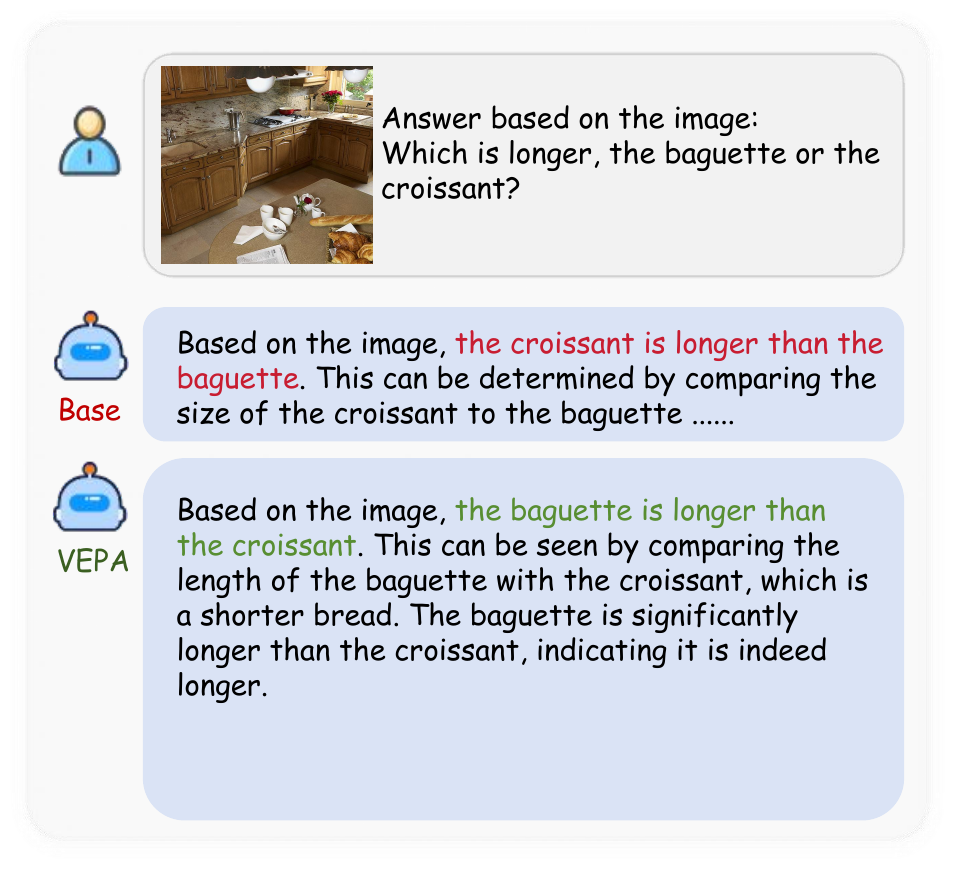}
    \captionof{figure}{Case 5 on Comparison}
    \label{fig:case5-comparison}
  \end{minipage}
  \hfill
  \begin{minipage}[t]{0.49\linewidth}
    \centering
    \includegraphics[width=\linewidth]{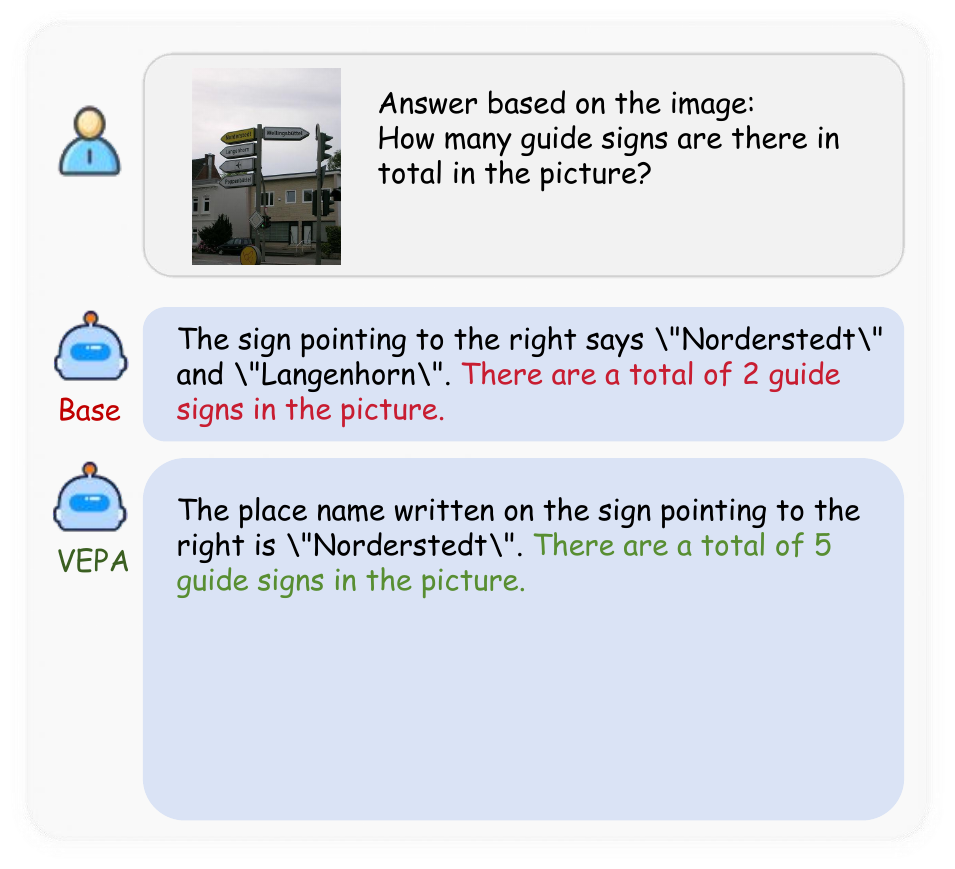}
    \captionof{figure}{Case 6 on Counting}
    \label{fig:case6-counting}
  \end{minipage}
\end{figure}

We present six qualitative examples in Figures~\ref{fig:case1-counting}--\ref{fig:case6-counting} to probe \emph{whether VEPA activates perception}. To minimize confounds, the \textbf{base} model in this subsection is the \emph{purely pre-trained} Qwen2-VL-2B checkpoint (i.e., \emph{before} any SFT or preference post-training). We then compare it with the corresponding \textbf{VEPA} model obtained by applying only the intermediate VEPA stage on top of this base. Importantly, we do \emph{not} enforce any extra response format: the model is simply asked to answer each question with the default prompt, without additional instructions that would explicitly demand “look at the image,” “explain,” or “provide evidence.” This design allows us to attribute qualitative differences primarily to changes in \emph{visual processing} induced by VEPA, instead of compliance with explicit instruction templates.

Under this controlled setting, we observe a consistent shift in the information the model chooses to attend to and verbalize. The base model frequently falls back to under-specified, generic, or prior-driven responses, especially for \emph{counting} and \emph{comparison}, where success requires enumerating multiple entities and tracking their relations. Its outputs often omit critical perceptual details (e.g., the exact number of instances, distinguishing attributes, or the relevant subset defined by the question), making the answers difficult to justify from the image. By contrast, the VEPA model is more likely to \emph{spontaneously} surface question-relevant visual facts—explicitly enumerating objects, mentioning discriminative attributes, reading visible text when needed, and describing spatial/relational cues that enable the downstream reasoning step. Notably, this behavior emerges \emph{without} being instructed to produce explanations, suggesting that VEPA increases the model’s tendency to consult and extract task-relevant visual evidence during generation.

These cases therefore provide qualitative support for our mechanism-level claim. Since both models share the same architecture and the same pretraining, and since we do not add instruction constraints at inference time, the observed differences are most naturally explained by VEPA strengthening the model’s internal reliance on perceptual signals. In other words, VEPA appears to \emph{activate} (and make accessible) visual perception for decision making, rather than simply inducing a different style of instruction-following responses.

\subsection{Training Dynamics}
\label{appendix:training}

This appendix summarizes additional training-time diagnostics used to monitor the stability and convergence of VEPA optimization.
Beyond the downstream evaluations in the main paper, we track two classes of signals during training:
(i) \emph{rollout statistics}, measured by the mean generated response length of the auxiliary model (i.e., the auxiliary model's output length under the $(q,e)$ prompt), and
(ii) \emph{optimization signals}, measured by the sequence-level reward used for sufficiency-driven GRPO.
Since both quantities are computed from stochastic rollouts, the curves can be noisy; we therefore visualize the raw traces together with an exponential moving average (EMA, weight $0.85$) to highlight the underlying trend.

\begin{figure}[!htbp]
    \centering
    \includegraphics[width=0.9\textwidth]{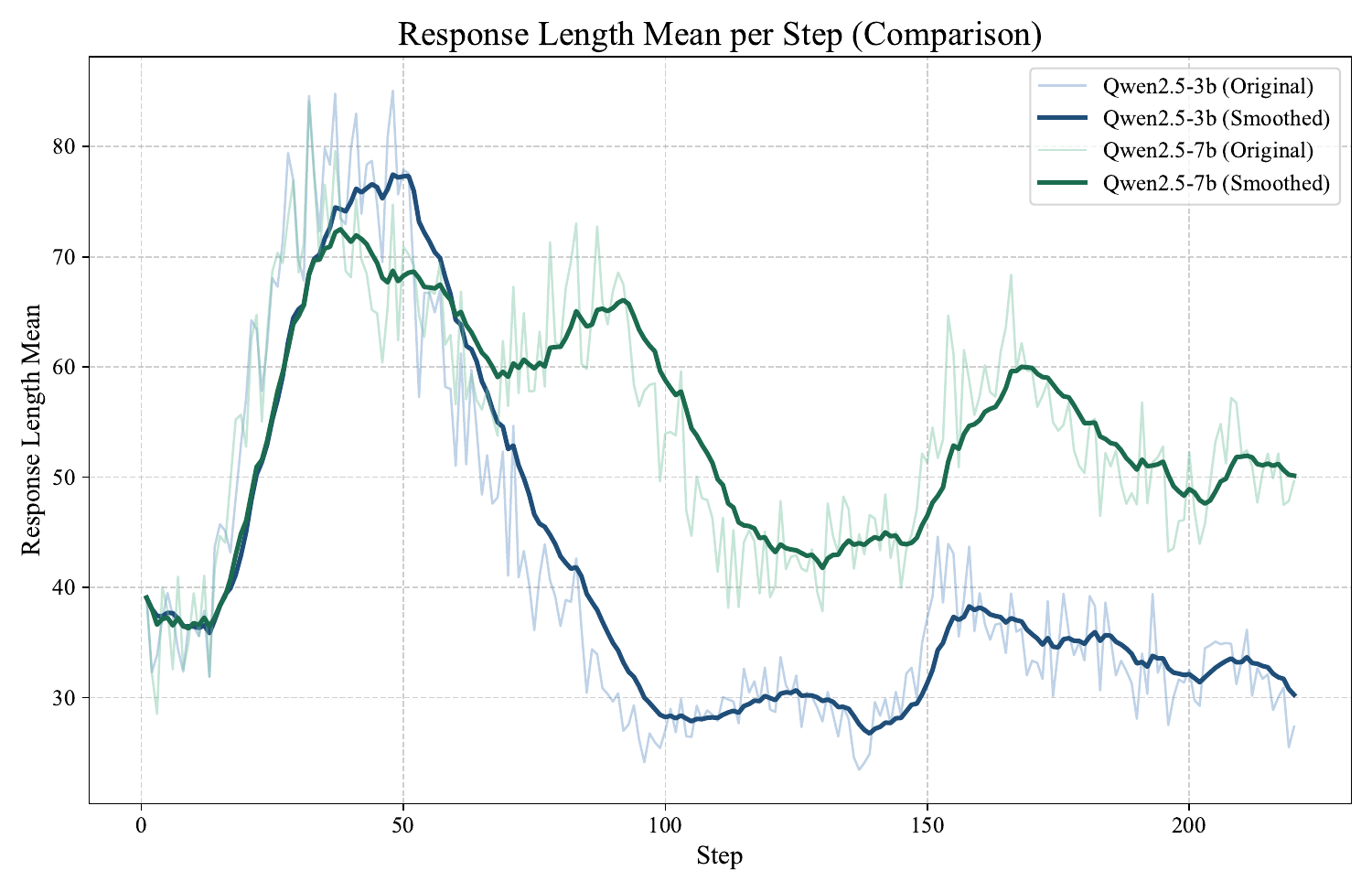}
    \caption{\textbf{Evaluation Metrics of Auxiliary Model Training.}
    The figure presents the evolution of the mean response length (y-axis) over training steps (x-axis).
    The blue and green lines correspond to the two distinct auxiliary model settings evaluated in this experiment.
    For each setting, the solid darker lines indicate the smoothed values (exponential moving average, weight 0.85), while the faint background lines represent the raw recorded data points.}
    \label{fig:traing_reponse_lentgh}
\end{figure}

\paragraph{Rollout length.}
Figure~\ref{fig:traing_reponse_lentgh} reports the evolution of the mean response length over training steps for two auxiliary model settings.
This metric serves as a lightweight diagnostic of rollout behavior, helping verify that training does not exhibit degenerate length collapse or uncontrolled length inflation while optimizing the evidence policy.

\begin{figure}[!htbp]
  \centering
  \begin{minipage}[t]{0.9\linewidth}
    \centering
    \includegraphics[width=\linewidth]{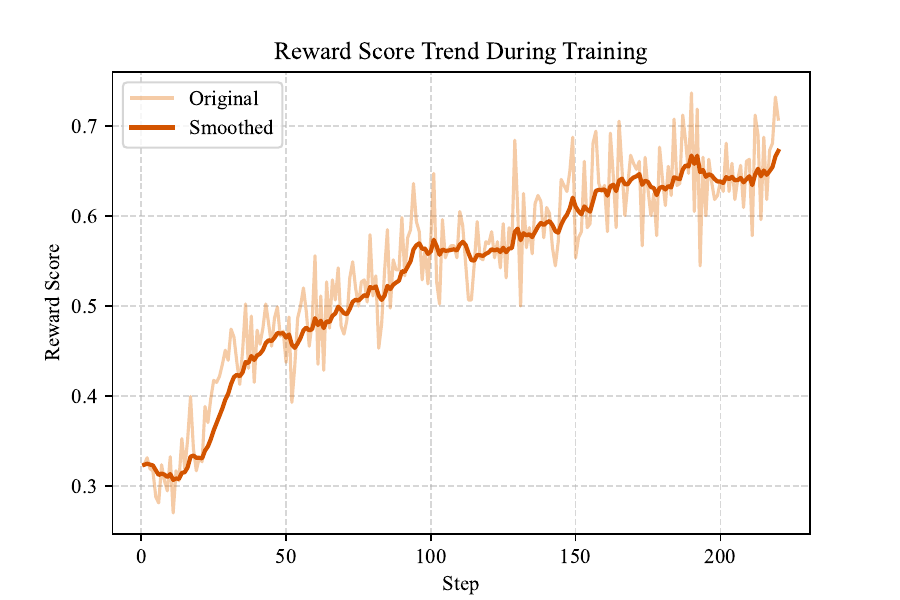}
    \caption{Reward Score Trend During Training (5k-VEPA-3b-aux-critic).}
    \label{fig:training-5k-VEPA-3b}
  \end{minipage}

  \vspace{6pt}

  \begin{minipage}[t]{0.9\linewidth}
    \centering
    \includegraphics[width=\linewidth]{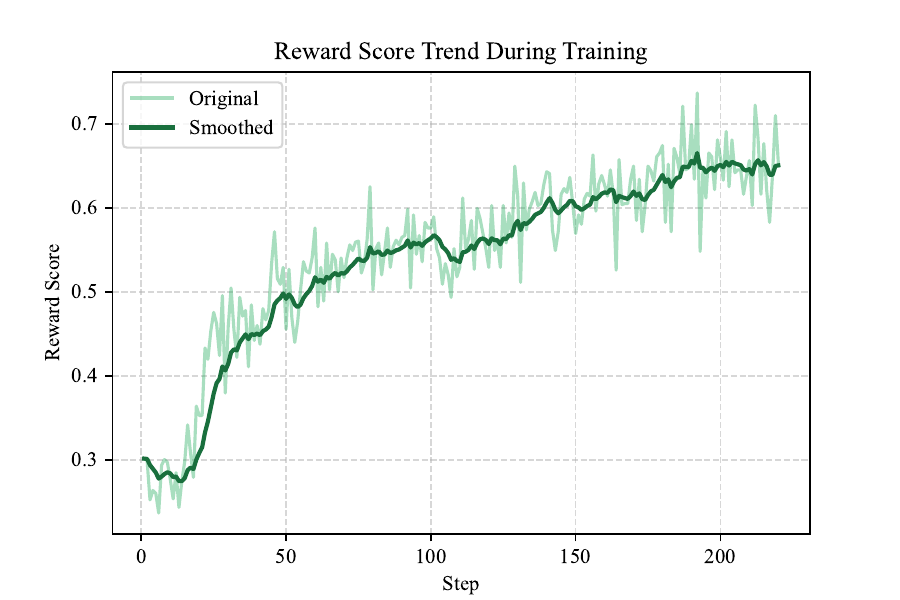}
    \caption{Reward Score Trend During Training (5k-VEPA-7b-aux-critic).}
    \label{fig:training-5k-VEPA-7b}
  \end{minipage}
\end{figure}

\begin{figure}[!htbp]
  \centering
  \begin{minipage}[t]{0.9\linewidth}
    \centering
    \includegraphics[width=\linewidth]{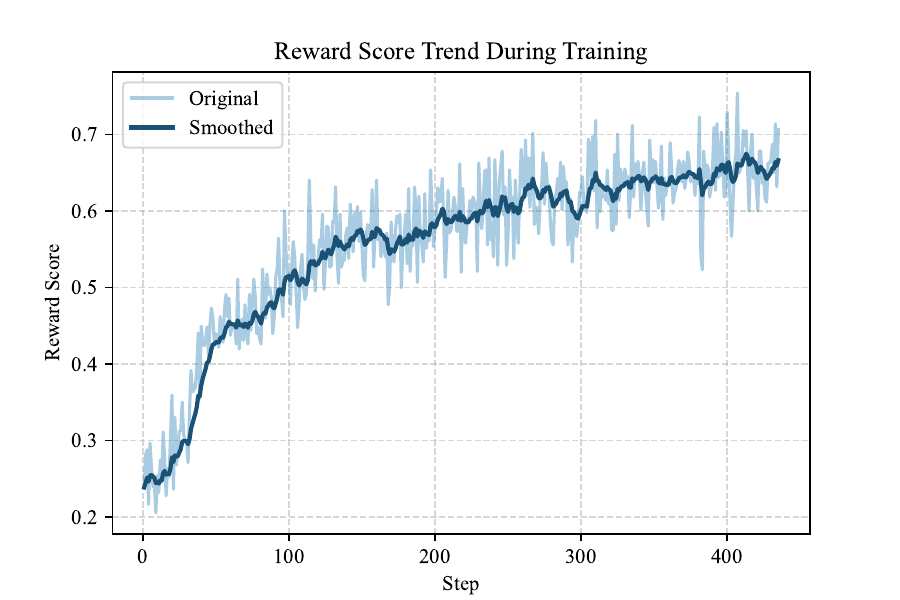}
    \caption{Reward Score Trend During Training (10k-VEPA-7b-aux-critic).}
    \label{fig:training-10k-VEPA-7b}
  \end{minipage}
\end{figure}

\paragraph{Reward dynamics.}
Figures~\ref{fig:training-5k-VEPA-3b}--\ref{fig:training-10k-VEPA-7b} plot the reward trajectories for different VEPA configurations.
Each figure shows both the per-step reward (raw) and its smoothed counterpart (EMA).
Overall, the reward trends provide an at-a-glance view of optimization progress and stability under sequence-level reinforcement learning for free-form evidence generation.

\label{appendix:training}

\newpage
  \begin{table}[!htbp]
    \centering
    \caption{\textbf{Reproducibility experiments across datasets.}
    We report mean $\pm$ SD over 3 independent runs for both \textbf{SFT} and \textbf{VEPA+SFT} settings.}
    \label{tab:reproducibility}
    \small
    \setlength{\tabcolsep}{10pt}
    \renewcommand{\arraystretch}{1.18}
    \begin{tabular}{@{}llccc@{}}
      \toprule
      \textbf{Dataset} & \textbf{Benchmark} & \textbf{SFT (mean$\pm$SD)} & \textbf{VEPA+SFT (mean$\pm$SD)} & \textbf{$\Delta$} \\
      \midrule
      GQA & A-OKVQA & 42.02$\pm$0.03 & 42.77$\pm$0.58 & +0.75 \\
          & GQA    & 63.86$\pm$0.84 & 64.61$\pm$0.10 & +0.75 \\
      \midrule
      A-OKVQA & A-OKVQA & 59.07$\pm$0.74 & 67.53$\pm$0.85 & +8.46 \\
              & ChartQA & 51.13$\pm$0.71 & 59.12$\pm$0.21 & +7.99 \\
              & TextVQA & 62.05$\pm$0.70 & 66.37$\pm$0.37 & +4.32 \\
      \midrule
      Mixed & A-OKVQA & 62.46$\pm$0.08 & 62.70$\pm$0.05 & +0.24 \\
            & ChartQA & 70.06$\pm$0.27 & 70.53$\pm$0.19 & +0.47 \\
            & TextVQA & 79.63$\pm$0.19 & 79.96$\pm$0.22 & +0.33 \\
      \bottomrule
    \end{tabular}
  \end{table}
\clearpage
\newpage

\end{document}